\documentclass[conference]{IEEEtran}
\IEEEoverridecommandlockouts
\usepackage{cite}
\usepackage{amsmath,amssymb,amsfonts}
\usepackage{algorithmic}
\usepackage{algorithm}
\usepackage{algorithmic}
\usepackage{multirow}
\usepackage{graphicx}
\usepackage{textcomp}
\usepackage{xcolor}
\usepackage{booktabs}
\usepackage{graphicx}
\usepackage{subfigure}
\usepackage{flushend}
\AtBeginDocument{%
  }
\begin{document}

\title{AdapterFL: Adaptive Heterogeneous Federated Learning for Resource-constrained Mobile Computing Systems
}


\author{
\IEEEauthorblockN{Ruixuan Liu$^1$, Ming Hu$^{2*}$\thanks{* Corresponding authors: Ming Hu (hu.ming.work@gmail.com) and Mingsong Chen (mschen@sei.ecnu.edu.cn)}, Zeke Xia$^1$, Jun Xia$^3$, Pengyu Zhang$^1$, Yihao Huang$^2$, Yang Liu$^2$, Mingsong Chen$^{1*}$}
\IEEEauthorblockA{
\textit{$^1$MoE Engineering
Research Center of SW/HW Co-Design Tech. and App., East China Normal University, China}\\
\textit{$^2$School of Computer Science and Engineering, Nanyang Technological University, Singapore} \\
 \textit{$^3$Computer Science and Engineering, University of Notre Dame, USA}
}
}

\maketitle

\begin{abstract}
Federated Learning (FL) enables collaborative learning of large-scale distributed clients without data sharing. However, due to the disparity of computing resources among massive mobile computing devices, the performance of traditional homogeneous model-based Federated Learning (FL) is seriously limited.
On the one hand, to achieve model training in all the diverse clients, mobile computing systems can only use small low-performance models for collaborative learning. On the other hand, devices with high computing resources cannot train a high-performance large model with their insufficient raw data.
To address the resource-constrained problem in mobile computing systems, we present a novel heterogeneous FL approach named AdapterFL, which uses a model reassemble strategy to facilitate collaborative training of massive heterogeneous mobile devices adaptively.
Specifically, we select multiple candidate heterogeneous models based on the computing performance of massive mobile devices and then divide each heterogeneous model into two partitions. 
By reassembling the partitions, we can generate models with varied sizes that are combined by the partial parameters of the large model with the partial parameters of the small model.
Using these reassembled models for FL training, we can train the partial parameters of the large model using low-performance devices.
In this way, we can alleviate performance degradation in large models due to resource constraints.
The experimental results show that AdapterFL can achieve up to 12\% accuracy improvement compared to the state-of-the-art heterogeneous federated learning methods in resource-constrained scenarios.

\end{abstract}

\section{Introduction}
\label{Introduction}
With the improvement of the capability of mobile Artificial Intelligence (AI) computing chips and the relevant hardware of mobile devices, AI applications based on mobile computing are becoming a trend~\cite{oort}.
Although Web of Things~(WoT)~\cite{advances,2016challenges,asynchronous} technologies enable efficient communication and data transmission between different devices, the data collected by mobile devices such as mobile phones and wearable devices usually involves user privacy.
Due to concerns about the risk of privacy leakage, the traditional centralized paradigm, which collects all the data in a central cloud, is challenging to satisfy mobile computing requirements.

As a well-known distributed machine learning paradigm, Federated Learning (FL)~\cite{FedAvg} enables various clients to train a global AI model without data sharing collaboratively.
Due to the advantage of privacy protection, FL has been widely used in mobile computing and WoT applications, such as real-time systems~\cite{li2019smartpc,hu2022gitfl}, IoT systems~\cite{zhang2020efficient,aiotml}, and autonomous driving systems~\cite{autonomous}.
In each FL round, the cloud server dispatches a global model to multiple clients for local training. Each client uses its raw data to train the received model and then uploads the trained model to the cloud server.
By aggregating uploaded models, the cloud server generates a new global model for local training of the next FL round.
In this way, the cloud server achieves model training without leaking privacy.

Although FL is promising in privacy protection due to using the same global model for local training, there still exist three challenges in mobile computing and WoT systems.
The first challenge is the heterogeneity of the devices in the WoT systems.
Since the computing capability of hardware resources (e.g., CUP and GPUs) in WoT devices is quite different~\cite{DBLP:journals/spm/LiSTS20,asynchronous,hierarchical}, as the bucket effect reveals, for traditional FL, the selection of the global model depends on the lowest-performance device.
In other words, the cloud server has to select the small low-performance model as the global model, which causes i) the hardware resources of high-performance devices not to be fully utilized and ii) high-performance devices to be deployed in a low-performance model with poor inference accuracy.
The second challenge is that the data of each device is limited.
Due to the data limitation of each device, it is difficult to train a usable high-performance model on a small group of high-performance devices.
The third challenge is that device data are typically non-IID (Independent and Identically Distributed)~\cite{non-iid}. Since WoT devices are deployed in different physical environments, the distributions of their collected data are affected by environments and user preferences, resulting in the problem of ``client drift''~\cite{scaffold} and causing the inference accuracy degradation of the aggregated global model.

To improve the performance of FL, existing methods can be classified into two categories, i.e., homogeneous methods and heterogeneous methods.
Homogeneous FL methods~\cite{FedAvg, adaptive, scaffold, FedGen} still use the same model as the global model for local training. This goal aims to use a wisely model training mechanism~\cite{scaffold, FedGen}, device selection mechanism~\cite{criticalfl, sample, fededge,fedltd}, or a data processing mechanism~\cite{smartphonedata, healthdataanalytics, store} to improve the inference performance of the global model.
Although homogeneous FL methods can alleviate performance degradation caused by non-IID data, their performance is still limited due to existing low-performance devices.
The heterogeneous methods~\cite{HeteroFL, InclusiveFL, Split-Mix, depthfl,hierarchyfl} attempt to dispatch multiple heterogeneous models to different devices for local training. In this way, high-performance devices can train a larger model rather than a low-performance small model. The cloud server can enable knowledge transfer among heterogeneous models using hyper-networks~\cite{HeteroFL, depthfl} or knowledge distillation technologies~\cite{distilling, FedDF}.
Although these heterogeneous methods can improve resource utilization, it is usually challenging to ensure that all the models achieve usable performance due to resource constraints, such as limited data on high-performance devices. 
In addition, existing heterogeneous methods usually rely on a specific structure of the hyper-network without considering the heterogeneity of device hardware. 
Specifically, different hardware architectures result in different processing capabilities for different model structures~\cite{efficient, eie,eyeriss}, such as convolutional layers and fully connected layers.
Therefore, \textit{how to improve FL performance in resource-constrained scenarios is a serious challenge for AI applications in mobile computing and WoT systems.}

Typically, although heterogeneous models have different structures, these models can be divided into multiple similar function modules, such as feature extraction and classification modules.
Intuitively, if modules with the same functions in the large-size model can be grafted into the small-size model, we can attempt to use low-performance devices to train partial model parameters. In this way, the partial parameters of the large-size model can be trained more adequately, and then their inference performance can be improved.
Many recent works~\cite{zhang2021understanding, transfer, transferable,transferlearning} have observed that early layers of a network tend to capture low-level, general features, while as the network becomes deeper, the features become more abstract and task-specific.
Inspired by the above observation, the heterogeneous models can be divided into two blocks, i.e.,  a feature-extraction block and a device-adaptation block.
By grafting the device-adaptation block from different heterogeneous models to the same feature-extraction block, we can generate multiple reassembled models, where the feature-extraction block can be trained by all the devices to extract the general features, and the device-adaptation block can be selected according to the hardware resource of the devices.
All the device-adaptation blocks are used to perform specific tasks according to the extracted features.

Based on the motivation above, we propose AdapterFL, a novel heterogeneous federated learning framework that implements collaborative training between heterogeneous models through model partition and reassembly. 
In AdpaterFL, the cloud server selects multiple heterogeneous prototype models according to the hardware resource of the devices and divides each prototype model into two blocks, i.e., the feature-extraction block and the device-adaptation block, respectively.
The cloud server then reassembles these blocks into a group of models with the same feature-extraction block.
These reassembled models are dispatched to local devices for local training in each FL training round.
When the model aggregation process is performed, the cloud server aggregates blocks with the same structure.
In this way, AdapterFL can enable FL to adaptively select heterogeneous models according to the hardware resource of the devices.
The main contributions of our paper are as follows:

\begin{itemize}
\item We propose AdapterFL, a novel heterogeneous FL framework that allows adaptively heterogeneous model selection based on the hardware resource of the devices.
\item We present a model reassembling mechanism to generate multiple heterogeneous models for FL training, where each model consists of a homogeneous feature-extract block and a heterogeneous device-adaptation block.
\item We conducted extensive empirical evaluations on three well-known datasets and various heterogeneous models to demonstrate the effectiveness of our AdapterFL approach.


\end{itemize}




\section{Related Work}
\label{Related Work}

\textbf{Heterogeneous Federated Learning.} Basic FL methods are based on the assumption that all clients have sufficient resources. However, when clients' resources are constrained, these methods are no longer applicable. So far, many FL methods for this problem have been proposed. Some existing works~\cite{FedMD, pervasivefl, HetComp, FedET, DS-FL} address this problem with heterogeneous models based on knowledge distillation~\cite{distilling}. For example, FedMD~\cite{FedMD} enhances the performance of models by distilling from the ensemble of predictions of heterogeneous local models, and HetComp~\cite{HetComp} reduces the huge inference costs while retaining high accuracy. However, these distillation-based methods require a public proxy dataset, which is often impractical. In addition, some works prune the global model into heterogeneous sub-models to solve the problem of model heterogeneity. For example, HereroFL~\cite{HeteroFL} and Split-Mix~\cite{Split-Mix} prune the global model by variant widths for clients with different resources and aggregate overlapping parameters of sub-models to share knowledge. However, they may have performance limitations since pruning the model by widths will destroy the model structure and cause parameter mismatch during aggregation. By pruning the model by variant depths, DepthFL~\cite{depthfl} circumvents this problem and further improves performance through self-distillation on clients. Other related methods, such as InclusiveFL~\cite{InclusiveFL} and FlexiFed~\cite{Flexifed}, improve the accuracy by aggregating heterogeneous models' common parts. However, these methods are limited by the overall architecture of models and must be applied to heterogeneous models with similar architecture. Our method is to solve the problem of model heterogeneity through model partition and reassembly, which is not limited by model architecture.

\textbf{Knowledge Transfer.}
Knowledge transfer~\cite{transferable,transferlearning} aims at transferring the knowledge learned by a model in a domain across others. There are a lot of existing knowledge transfer works~\cite{distilling, delta, yang2020transfer, Dery}.
A typical method in knowledge is to split the model into a lower (common) part and a higher (specific) part so that the knowledge learned by the former about an ML task can be shared with other ML tasks~\cite{transfer, Flexifed}. 
DeRy~\cite{Dery} is a novel knowledge-transfer method that proposes a model partition and reassembly method for pre-trained models. DeRy leverages the representations' similarity to quantify the distance among neural networks. It divides each neural network into several building blocks according to the similarity so that building blocks in the same position from different models can play similar roles and extract common features in the neural network. Then, this method can rearrange the building blocks from various neural networks in positional order to reassemble new models. AdpaterFL leverages the model partition method to divide the prototype model into two blocks, the lower feature-extraction block and the higher device-adaptation block, and rearranges them to reassemble more new models. The reassembled models with the same feature-extraction block can extract general features to achieve the purpose of knowledge transfer. Therefore, they can serve as a group of models for the problem of model heterogeneity in the resource-constrained scenario.

\section{PRELIMINARIES}
\label{PRELIMINARIES}

\begin{figure*}[ht]
\centering
\includegraphics[width=0.88\textwidth]{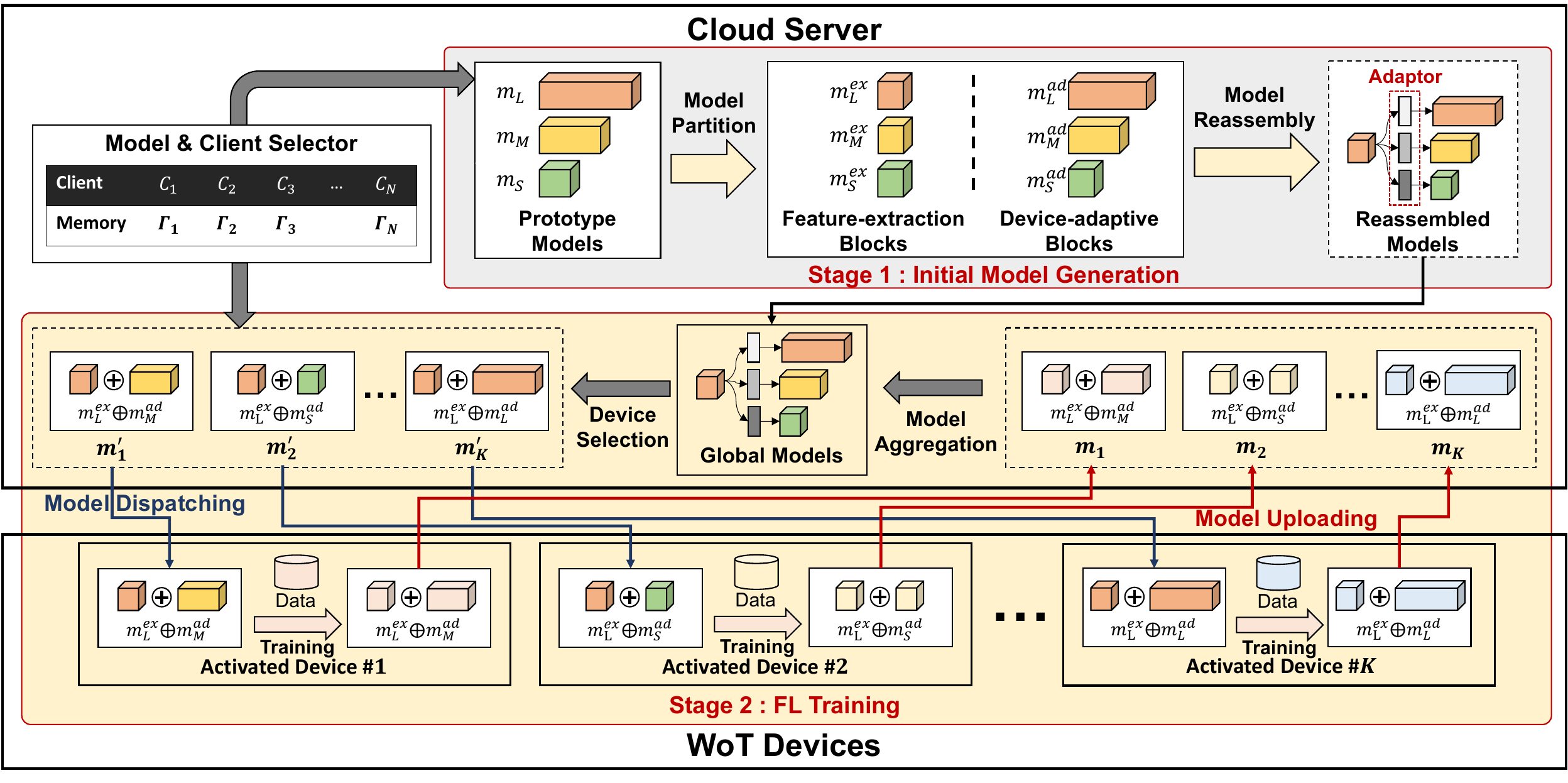} 
\caption{The framework and workflow of AdapterFL.}
\vspace{-0.1in}
\label{fig:framework}
\end{figure*}

In the general FL framework~\cite{FedAvg}, there are $K$ activated clients at each round. Each client has its private local dataset $D_k$ drawn from the distribution $P_k(x,y)$, where $k\in \left \{ 1,\cdots, K  \right \} $, $x$ and $y$ denote the input features and corresponding class labels respectively. The local loss function of each client is as follows:
\begin{equation}
\label{eq1}
	L_k(w_k) = \frac{1}{|D_k|} \sum_{i=1}^{|D_k|} \ell
\end{equation}
where $|D_k|$ is the number of instances in $|D_k|$, $(x_i,y_i)\in D_k$, $w_k$ is the parameters of the local model on $k^{th}$ client, and $\ell$ is the general loss function of any supervised learning task (e.g., the cross-entropy loss). The global objective function of the framework is:
\begin{equation}
\label{eq2}
	\underset{\bar{w}}{argmin}~L(\bar{w}) = \sum_{k=1}^{K}\frac{|D_k|}{N} L_k(w_k)
\end{equation}
where $N$ is the sum of $|D_k|$, $k\in \left \{ 1,\cdots, K  \right \} $, and $\bar{w}$ is the parameters of the global model. 

However, in the real-world scenario, some clients in FL have insufficient resources due to their poor computing capacity and memory. Since applying the same model to all clients in this scenario is impractical, the general FL optimization function will no longer apply to this scenario. Correctly dividing models can help us find model blocks with similar functions among heterogeneous models, which greatly contributes to solving the problem of model heterogeneity. In the model partition~\cite{Dery}, models are partitioned according to features similarity computed by the input-output similarity function $S(B, B') = s(B(x), B'(x')) + s(x,x')$, where $B(x)$ is the output feature of model sub-block $B$ with input $x$ and $s(x,y)$ is a method to measure the similarity between $x$ and $y$, such as centered kernel alignment (CKA)~\cite{cka}, canonical correlation analysis (CCA)~\cite{cca}. The optimization function that divides each model $M_i$ in the model zoo $M$ into two parts $B_i^0$ and $B_i^1$ is:
\begin{equation}
\footnotesize
\begin{split}
\label{eq3}
    \left \{ \left \langle  B_i^0,B_i^1 \right \rangle \right \} _{i=1}^{|M|}  &= 
    \underset{f\in\{ 1,|L|\}}{argmax}\sum_{i=1}^{|M|}\left [ S(B_i^{0,f},B_{an}^{0}) + S(B_i^{1,f},B_{an}^{1}) \right ] \\
    &s.t. \ \  B_i^0 \circ   B_i^1 = M_i, B_i^0  \cap   B_i^1 = \phi  .  
\end{split}
\end{equation}
where $B_{an}$ is the anchor block with the maximum summed similarity with other blocks, $|M|$ and $|L|$ indicate the number of models and layers of the corresponding model, respectively. $B_i^{0,f}$ and $B_i^{1,f}$ represent respectively the block containing the $0^{th}$ to the $f^{th}$ layer and the $f^{th}$ to the last layer of model $M_i$. We can find the optimal partition for these models by solving the function above.


\section{Our AdapterFL Approach
}
\label{METHODOLOGY}
\subsection{Overview
}
To address the challenges of resource constraints in mobile computing systems, we propose a novel heterogeneous federated learning framework named \textit{AdapterFL}.
Figure~\ref{fig:framework} presents the framework and workflow of our AdapterFL approach, which consists of a cloud server and multiple WoT devices. 
To adapt to various devices with limited resources, AdapterFL includes a model \& client selector, which collects hardware information from WoT devices to i) select prototype models and ii) assign heterogeneous global models for each device.
The workflow of AdapterFL consists of two stages, i.e., the initial model generation stage and the FL training stage respectively.
The initial model generation stage selects multiple prototype models according to the different hardware resources of WoT devices and reassembles these prototype models to generate multiple heterogeneous global models for FL training.
The FL training stage dispatches the generated heterogeneous models to WoT devices for local training according to their hardware resources and then aggregates the trained models to update global models.


As shown in Figure~\ref{fig:framework}, specifically, the initial model generation stage has two steps, i.e., model partition and model reassembly, respectively. 
In the model partition step, the cloud server divides each prototype model into two blocks, i.e., the feature-extraction block and the device-adaptation block, respectively.
In Figure~\ref{fig:framework}, there are three heterogeneous prototype models, i.e., $m_L$, $m_M$, and $m_S$, which denote the large model, the medium model, and the small model, respectively.
Typically, the size difference of feature-extraction blocks is much smaller than that of device-adaptation blocks.
In the model reassembly step, the cloud server selects a feature-extraction block from the divided feature-extraction blocks in the first step and reassembles it with all the device-adaptation blocks to generate multiple initial heterogeneous global models.
In Figure~\ref{fig:framework}, we select $m^{ex}_L$ as the feature-extraction block of all the global models.
Since the structures of the device-adaptation blocks are different, for each device-adaptation block, AdapterFL uses an adapter to connect the feature-extraction block.
This way, the cloud server can generate multiple heterogeneous global models with the same feature-extraction block.

The FL training stage consists of multiple FL training rounds, where each FL training round has five steps, i.e., device selection, model dispatching, local training, model uploading, and mode aggregation.
In the device selection step, the cloud server selects $K$ activated devices for model training according to their hardware resources and assigns an appropriate global model for each activated device.
In the model dispatching step, the cloud server dispatches global models to corresponding devices.
In the local training step, activated devices use their raw data to train their deceived global model.
In the model uploading step, each activated device uploads its trained model to the cloud server.
In the model aggregation step, the cloud server aggregates the corresponding blocks of all the received models to update the global models.

\subsection{Initial Model Generation}
Since the hardware and data resources of WoT devices are seriously limited in mobile computing systems, the goal of our AdapterFL is to generate multiple heterogeneous models to adapt WoT devices with various hardware configurations and can be sufficiently trained with limited data.
AdapterFL divided each prototype model into two blocks, i.e., the feature-extraction block and the device-adaptation block. By reassembling a feature-extraction block with all the device-adaptation blocks, the cloud server generates multiple initial global models.
In this way, the cloud server can select a more suitable global model according to the hardware resource of the target device. Since all the reassembled models have the same feature-extraction block, such a block can be trained by all the devices and then alleviate inadequate training of large-size models.



\subsubsection{Model Partition}
Inspired by DeRy~\cite{Dery}, which observed that different pre-trained neural networks can be divided into multiple similar blocks by calculating the similarity of their functional features and such blocks can be reassembled into new usable models, we use CKA~\cite{cka} as the metric to divide each prototype model.
Specifically, for each prototype model $m_i \in M$, we can divide it into two blocks: the feature-extraction block $m_{i}^{ex}$ and the device-adaptation block $m_{i}^{ad}$ by Equation~\ref{eq3}:
\begin{equation}
\label{eq4}
	m_{i} = \left \langle m_{i}^{ex}, m_{i}^{ad}  \right \rangle 
\end{equation} 
After the partition, each prototype model is divided into two blocks based on functional similarity. Then, we can use these blocks to reassemble new models, which will be dispatched to clients for local training. 

\subsubsection{Model Reassembling}
In the model reassembling step, the cloud server selects a feature-extraction block from the divided feature-extraction blocks in the model partition step.
Note that since the sizes of different divided feature-extraction blocks are similar, to generate higher-performance models, the cloud server prefers to select the feature-extraction block divided from a large-size model for model reassembling.
By combining the selected feature-extraction block with all device-adaptation blocks, the cloud server can generate a group of new reassembled models. 
However, since the feature-extraction block and the target device-adaptation block may come from different models, their feature dimensions are mismatched.
To achieve reassembling between heterogeneous blocks, the cloud server generates an adapter for each reassembly model to align the dimensions.
In our framework, the adapter contains two convolutional layers, which will be attached to the feature-extraction block and the device-adaptation block, respectively. By the connecting of an adapter, a feature-extraction block $m_{i}^{ex}$ and a device-adaptation block $m_j^{ad}$,  a reassemble model $m_{i,j}$ can be formed as follows:

\begin{equation}
\label{eq5}
	m_{i,j} = m_{i}^{ex} \oplus  m_{j}^{ad}  = \left \langle (m_{i}^{ex} \circ  \alpha ^ 0 ) , (\alpha ^1 \circ m_{j}^{ad} ) \right \rangle
\end{equation} 
where $m_{i}^{ex}$ and $m_{j}^{ad} $ represent the $i^{th}$ feature-extraction block and the $j^{th}$ device-adaptation block, respectively.
$\alpha ^ 0$ and $\alpha ^ 1$ represent the first and second convolutional layers of the adapter, respectively. 
\subsection{Device Selection and Model Dispatching}
Since the sizes of different reassembled models are quite different, our \textit{Model \& Client Selector} maintains a table to record the hardware resource of each device. In each FL training round, the selector dispatches these reassembled models to activated clients when $|m_{i,j}| \leq \Gamma_k $, where $|m_{i,j}|$ is the required memory for the deployment of $m_{i,j}$ and $\Gamma_k$ denotes the available memory of the $k^{th}$ device. 
To achieve sufficient training of the global models, our device selection strategy ensures that each heterogeneous global model is dispatched to at least one device for training in each FL training round.
Moreover, the number of devices to dispatch for a global model at each round is determined according to the number of suitable devices for this global model.

\subsection{Block-based Aggregation}
Once reassembled models are locally trained on clients, their parameters are uploaded back to the server. Since these models trained on clients are heterogeneous, it is impossible to aggregate all models according to traditional methods, e.g., Equation~\ref{eq2}. However, these models contain the same feature-extraction block, which enables us to use the model aggregation methods based on building blocks. We separately calculate the parameters of the feature-extraction block $w_i^{ex}$ and the device-adaptation block $w_j^{ad}$ to obtain the parameters of the global reassembled model $m_{i,j} \in G_i$. The parameters $w_i^{ex}$ are obtained by aggregating the parameters of feature-extraction blocks from all the clients:
\begin{equation}
\label{eq6}
	\bar{w}_{i}^{ex}  = \frac{1}{K} \sum_{k=1}^{ K} w_{i,k}^{ex}
\end{equation}
where $w_{i,k}^{ex}$ is the parameters of the feature-extraction block in the model from the $k^{th}$ activated client. However, the parameters of the device-adaptation block $w_{j}^{ad}$  are obtained by only aggregating device-adaptation blocks from those models with the common structure: 
\begin{equation}
\label{eq7}
	\bar{w}_{j}^{ad}  = \frac{1}{|C|} \sum_{k \in C}^{C} w_{j,k}^{ad}
\end{equation}
where $c$ is the set of clients that train the model $m_{i,j}$. In simple terms, our method aggregates blocks with a common structure. Since shared blocks are aggregated from different heterogeneous models, they can share the knowledge between heterogeneous models by extracting the general features. 
Compared with the small model, the large model has stronger generalization and higher performance limits but requires more computing resources. The large model may suffer from insufficient model training problems in resource-constrained scenarios. However, in AdapterFL, since these heterogeneous models have the same lower feature-extraction block that can extract general features, we can perform knowledge sharing and migration between heterogeneous models and jointly improve overall performance through the joint aggregation of the lower block among heterogeneous models. In addition, generally speaking, the feature-extraction block of a large prototype model has stronger knowledge transfer and feature extraction capabilities than a small prototype model.


\subsection{Implementation of AdapterFL}
 Algorithm~\ref{alg:algorithm1} presents the implementation of our AdapterFL approach.
Lines 2-10 present the process of the initial model generation stage.
Lines 2-4 show the model partition step, which divides each prototype model $m_i$ into two blocks, i.e., $m_i^{ex}$ and $m_i^{ad}$. 
Lines 6-10 present the model reassembly and group these reassembly models into one group $G_i$ based on the feature-extraction block $m_i^{ex}$. 
Lines 12-25 present the process of the FL training stage.
Lines 13-14 present the device selection and model dispatching process. The cloud server contains a \textit{model \& client selector}, which records the model threshold $\Gamma_n$ of client $C_n$. It dispatches the model $m_{i,j}$ to client $C_k, s.t.  |m_{i,j}| < \Gamma_k$ where $|m_{i,j}|$ means the number of parameters of $m_{i,j}$. Lines 15-18 present the local training and model uploading.
Lines 20-24 present that the cloud servers aggregate the feature-extraction and device-adaptation blocks of reassembled models, respectively. Finally, we get the group $G_i$ as output, and this group contains three heterogeneous models with varied parameters.

\begin{algorithm}[h]
\caption{Implementation of AdapterFL}
\label{alg:algorithm1}
\begin{flushleft}
\textbf{Input}: 
i) $T$, the total training rounds; ii) $C$, the set of clients; iii) $r$, the ratio of client resource iv) $\Gamma_k$, the resources threshold of client $C_k$; v) $M = \{ m_S, m_M, m_L \}$, the prototype models \\
\textbf{Output}: $G_i$\\
\end{flushleft}
\begin{algorithmic}[1] 
\STATE \textit{/* Initial Model Generation */}
\FOR{each  $i \in \{S, M, L \}$}
\STATE Divide $m_i$ into two blocks $\left \langle m_{i}^{ex}, m_{i}^{ad}  \right \rangle$ via Equation~(\ref{eq3})
\ENDFOR
\STATE Randomly select a feature-extraction block $m_{i}^{ex}, i\in \{S, M, L \}$
\STATE $G_i = \phi$
\FOR{each $j \in \{S, M, L \}$}
\STATE Combine $m_i^{ex}$ and $m_j^{ad}$ to get $m_{i,j}$ via Equation~(\ref{eq5})
\STATE Add $m_{i,j}$ into $G_i$
\ENDFOR
\STATE \textit{/* FL Training */}
\FOR{each round $t = 1,...,T$}
\STATE Sample $K$ clients 
\STATE Allocate model $m_{i,j} \in G_i$ to $C_k$ $s.t. \ \ |m_{i,j}| < \Gamma_k$ with the ratio $r$ on the \textit{Model \& Client Selector } 
\FOR{\textbf{each client $C_k$ in parallel}}
\STATE Train $m^k_{i,j}$ with Equation~(\ref{eq1}) on private data
\STATE Upload $m^k_{i,j}$ to the cloud server
\ENDFOR
\STATE \textbf{Cloud Server}
\STATE Compute block $m_i^{ex}$ by aggregating via Equation~(\ref{eq6})
\FOR{each $j \in \{S, M, L \}$ }
\STATE Compute block $m_j^{ad}$ by aggregating via Equation~(\ref{eq7})
\ENDFOR
\STATE $m_{i,j} =  \left \langle m_{i}^{ex}, m_{j}^{ad}  \right \rangle \in G_i$
\ENDFOR
\STATE \textbf{return} $G_i$
\end{algorithmic}
\end{algorithm}
\vspace{-0.1in}

\section{Experiments}
\label{Experiments}

To evaluate the performance of AdapterFL, we conducted extensive experiments based on a variety  of 
well-known datasets and models.
All the experiment results were collected 
from a Ubuntu workstation equipped with an Intel i9 CPU, 32GB memory, and an NVIDIA RTX 3090Ti GPU. 
We implemented the AdapterFL framework 
on top of Pytroch (version 2.0.1). The following 
subsections aim to answer the following three Research Questions (RQs).


\textbf{RQ1 (Superiority of AdapterFL)}: What are the advantages of AdapterFL compared with state-of-the-art methods?

\textbf{RQ2 (Adaptivity of AdapterFL)}: What is the adaptivity of AdapterFL under different settings (e.g., client data distributions, datasets, DNN architectures)?

\textbf{RQ3 (Scalability of AdapterFL)}: What is the impact of different settings of clients on AdapterFL (e.g., ratios of resources-constrained clients, total number of clients)?

\subsection{Experimental Settings}
To ensure the fairness of the experiment, for each client, we set the batch size of local training to 50 and ran 5 epochs in each round.  For all the mentioned FL methods, we used Stochastic Gradient Descent (SGD)~\cite{overview} as the optimizer with a learning rate of 0.01, a learning rate decay of 0.998, a momentum of 0.5 and a weight decay of $1e^{-3}$ during local training.

\subsubsection{Dataset Settings}
We conducted experiments to investigate the performance of AdapterFL on three well-known datasets with both IID and non-IID scenarios, i.e., CIFAR-10, CIFAR-100, and TinyImagenet. For each dataset, we adopted the Dirichlet distribution~\cite{measuring} denoted by $p_c\sim Dir_k(\beta )$ to control the heterogeneity of client data, which $p_{c,k}$ is the ratio of data samples belonging to class c to client k and $Dir_k(\beta )$ is a Dirichlet distribution determined by $\beta$, where the smaller $\beta$ indicates the higher heterogeneity of client data. For all the datasets, we assumed that there are 100 clients in the FL architecture and selected $10\%$ of clients for local training and global aggregation at each round by default. 

\subsubsection{Device Heterogeneity setting}
To simulate heterogeneous device scenarios with limited resources, we classified 100 clients into three resource levels (small, medium, and large) corresponding to the three prototype models to simulate the device resource-constrained scenario. For all experiments, we set the client resource ratio of the 3 levels to 0.4-0.4-0.2 by default. This also aligns with the situation where clients with large computing capacity account for a relatively small proportion. In addition, we will also deliver these heterogeneous models according to this ratio.

\subsubsection{Model Setting}

By default, we used CNN~\cite{cnn} composed of 2 convolution layers, MobileNetV2~\cite{mobilenetv2} and ResNet18~\cite{resnet18} as three candidate models (small, medium, and large), respectively. To demonstrate the pervasiveness of our method, we used MobileNetV2, ResNet18, and Vgg16~\cite{Vgg} as another group of models with more parameters, as shown in Table~\ref{parameter}. For these candidate models, their upper limit of accuracy increases as the number of parameters of these models, but they require more computing resources. In the resources-constrained scenario, we assumed that each model could only be trained on clients that meet their computing resource requirements, such that a medium model could only run on clients of medium and large levels.
\begin{table}[h]
\vspace{-0.1in}
\caption{Parameters of candidate models}
\vspace{-0.1in}
\centering
\scriptsize
\begin{tabular}{@{}c|cccc@{}}
\toprule
Model      & CNN  & MobileNetV2 & ResNet18 & Vgg16\\ \midrule
Params (M) & 0.21 & 2.25        & 11.17   & 33.65 \\ \bottomrule
\end{tabular}%
\vspace{-0.15in}
\label{parameter}
\end{table}

\subsubsection{Baseline Methods Settings} 
We compared the inference accuracy of AdapterFL with three FL baseline methods. The following are the settings of these methods:
\begin{itemize}
\item FedAvg~\cite{FedAvg} is the most classic FL framework. In each round of training, the server dispatches the global model to activated clients for local training and aggregates uploaded local models to update the global model. In the resources-constrained scenarios, we conducted FedAvg with exclusive learning, which excluded clients whose resources can not afford the local training with the dispatched global model. In all tables, FedAvg (S, M, L) represents the results of training three candidate models separately with exclusive learning. 

\item FedDF~\cite{FedDF} is a KD-based FL framework. FedDF needs to maintain a global proxy dataset on the server. In each round of training, the server dispatches heterogeneous models to activated clients for local training. Then, the server uses the ensemble of these uploaded models to distill each model for knowledge sharing with the global dataset. In the experiment of FedDF, we select $1\%$ of the training data as the global proxy dataset on the server.
\item FedBase~\cite{FedAvg} is another baseline we defined, which trains reassembled models separately with exclusive learning and adapts FedAvg as the training framework. In our method, the model reassembly and the addition of the adaptor will change the model structure and increase the number of parameters slightly. Therefore, to eliminate the impact caused by model structure, we conduct experiments using FedBase with reassembled models. 

\end{itemize}

\begin{table}[h]
\caption{Comparison of reassembled models  on CIFAR-10}
\centering
\scriptsize
\begin{tabular}{@{}cc|ccc|c@{}}
\toprule
\multirow{2}{*}{Method}    & \multirow{2}{*}{Model} & \multicolumn{3}{c|}{Accuracy (\%)}                                 & \multirow{2}{*}{Params (M)} \\ \cmidrule(lr){3-5}
                         &     & IID         & $\beta=0.6$       & $\beta=0.3$       &       \\ \midrule
\multirow{3}{*}{FedAvg}    & S   & 67.84$\pm$0.13           & 64.66$\pm$0.29            & 63.16$\pm$0.50            & 0.21  \\
                           & M   & 58.28$\pm$0.18           & 60.11$\pm$0.52            & 48.21$\pm$1.23            & 2.25  \\
                           & L   & 53.21$\pm$0.19           & 48.56$\pm$0.24            & 45.95$\pm$0.41            & 11.17 \\ \midrule
\multirow{3}{*}{FedDF}     & S   & 67.68$\pm$0.15           & 66.73$\pm$0.18            & 63.34$\pm$0.42            & 0.21  \\
                           & M   & 63.56$\pm$0.11           & 64.66$\pm$0.34            & 60.99$\pm$0.66            & 2.25  \\
                           & L   & 58.60$\pm$0.22           & 59.06$\pm$0.37            & 49.83$\pm$0.93            & 11.17 \\ \midrule
\multirow{9}{*}{FedBase}   & S-S & 65.77$\pm$0.14           & 60.94$\pm$0.49            & 60.19$\pm$0.29            & 0.22  \\
                           & S-M & 62.01$\pm$0.10           & 59.07$\pm$0.53            & 49.49$\pm$1.15            & 2.20   \\
                           & S-L & 54.48$\pm$0.14           & 50.45$\pm$0.19            & 48.73$\pm$0.23            & 10.52 \\
                           & M-S & 68.36$\pm$0.16           & 65.19$\pm$0.27            & 61.13$\pm$0.33            & 0.28  \\
                           & M-M & 59.77$\pm$0.22           & 58.05$\pm$0.64            & 49.65$\pm$1.35            & 2.26  \\
                           & M-L & 48.83$\pm$0.20           & 46.62$\pm$0.23            & 43.58$\pm$0.18            & 10.58 \\
                           & L-S & 73.38$\pm$0.29           & 68.97$\pm$0.40            & 66.72$\pm$0.31            & 0.91  \\
                           & L-M & 68.39$\pm$0.21           & 64.77$\pm$0.45            & 57.72$\pm$1.28            & 2.89  \\
                           & L-L & 55.28$\pm$0.22           & 50.57$\pm$0.36            & 47.60$\pm$0.31            & 11.21 \\ \midrule
\multirow{9}{*}{AdapterFL} & S-S & 67.58$\pm$0.25           & 65.00$\pm$0.50            & 61.80$\pm$0.47            & 0.22  \\
                           & S-M & 63.89$\pm$0.21           & 60.06$\pm$0.42            & 52.45$\pm$2.53            & 2.20   \\
                           & S-L & 57.78$\pm$0.12           & 55.05$\pm$0.68            & 52.54$\pm$0.32            & 10.52 \\
                           & M-S & 67.75$\pm$0.21           & 65.37$\pm$0.28            & 60.06$\pm$0.53            & 0.28  \\
                           & M-M & 66.95$\pm$0.14           & 63.82$\pm$1.26            & 52.62$\pm$3.95            & 2.26  \\
                           & M-L & 64.22$\pm$0.10           & 61.72$\pm$0.46            & 55.93$\pm$0.56            & 10.58 \\
                           & L-S & 73.82$\pm$0.11           & 72.32$\pm$0.45            & \textbf{68.52$\pm$0.79}   & 0.91  \\
                           & L-M & \textbf{75.62$\pm$0.10}  & \textbf{72.34$\pm$0.82}   & 65.55$\pm$3.14            & 2.89  \\
                           & L-L & 69.54$\pm$0.08           & 66.61$\pm$0.69            & 64.66$\pm$0.74            & 11.21 \\ \bottomrule
\end{tabular}%
\vspace{-0.1in}
\label{accuarcy_reassembly_models}
\end{table}

\begin{table*}[ht]
\caption{Accuracy comparison on different datasets and distributions with L-group}
\centering
\scriptsize
\begin{tabular}{@{}cc|ccc|ccc|ccc|c@{}}
\toprule
\multirow{2}{*}{Method} &
  \multirow{2}{*}{Model} &
  \multicolumn{3}{c|}{CIFAR10} &
  \multicolumn{3}{c|}{CIFAR100} &
  \multicolumn{3}{c|}{TinyImageNet} &
  \multirow{2}{*}{Params (M)} \\ \cmidrule(lr){3-11}
 &
   &
  IID &
  $\beta=0.6$ &
  $\beta=0.3$ &
  IID &
  $\beta=0.6$ &
  $\beta=0.3$ &
  IID &
  $\beta=0.6$ &
  $\beta=0.3$ &
   \\ \midrule

\multirow{3}{*}{FedAvg} & S   & 67.84$\pm$0.13 & 64.66$\pm$0.29 & 63.16$\pm$0.50 & 23.41$\pm$0.09 & 25.20$\pm$0.20 & 25.15$\pm$0.12 & 11.87$\pm$0.10  & 13.95$\pm$0.20 & 15.38$\pm$0.14 & 0.21  \\
                        & M   & 58.28$\pm$0.18 & 60.11$\pm$0.52 & 48.21$\pm$1.23 & 21.81$\pm$0.21 & 24.68$\pm$0.14 & 23.36$\pm$0.20 & 23.23$\pm$0.27  & 23.67.0.22 & 22.72$\pm$0.25 & 2.25  \\
                        & L   & 53.21$\pm$0.19 & 48.56$\pm$0.24 & 45.95$\pm$0.41 & 20.90$\pm$0.20 & 19.59$\pm$0.43 & 18.26$\pm$0.22 & 27.57$\pm$0.23  & 26.25$\pm$0.43 & 24.69$\pm$0.44 & 11.17 \\ \midrule
\multirow{3}{*}{FedDF}  & S   & 67.68$\pm$0.15 & 66.73$\pm$0.18 & 63.34$\pm$0.42 & 26.94$\pm$0.13 & 29.56$\pm$0.17 & 28.84$\pm$0.28 & 19.69$\pm$0.22  & 20.73$\pm$0.27 & 21.47$\pm$0.24 & 0.21  \\
                        & M   & 63.56$\pm$0.11 & 64.66$\pm$0.34 & 60.99$\pm$0.66 & 25.44$\pm$0.16 & 25.90$\pm$0.24 & 25.57$\pm$0.11 & 30.51$\pm$0.18  & 29.99$\pm$0.17 & 30.16$\pm$0.24 & 2.25  \\
                        & L   & 58.60$\pm$0.22 & 59.06$\pm$0.37 & 49.83$\pm$0.93 & 22.44$\pm$0.23 & 21.92$\pm$0.26 & 21.08$\pm$0.53 & 26.24$\pm$00.30 & 24.18$\pm$0.25 & 22.29$\pm$0.40 & 11.17 \\ \midrule
\multirow{3}{*}{FedBase} &
  L-S &
  73.38$\pm$0.29 &
  68.97$\pm$0.40 &
  66.72$\pm$0.31 &
  35.80$\pm$0.22 &
  \textbf{38.84$\pm$0.24} &
  34.91$\pm$0.34 &
  22.67$\pm$0.22 &
  24$\pm$20$\pm$0.18 &
  23.96$\pm$0.23 &
  0.91 \\
                        & L-M & 68.39$\pm$0.21 & 64.77$\pm$0.45 & 57.72$\pm$1.28 & 24.18$\pm$0.17 & 24.80$\pm$0.28 & 23.73$\pm$0.24 & 19.70$\pm$0.23  & 21.81$\pm$0.28 & 20.77$\pm$0.22 & 2.89  \\
                        & L-L & 55.28$\pm$0.22 & 50.57$\pm$0.36 & 47.60$\pm$0.31 & 20.44$\pm$0.17 & 19.55$\pm$0.33 & 18.45$\pm$0.14 & 29.24$\pm$0.13  & 27.60$\pm$0.32 & 26.43$\pm$0.48 & 11.21 \\ \midrule
\multirow{3}{*}{AdapterFL} &
  L-S &
  73.82$\pm$0.11 &
  72.32$\pm$0.45 &
  \textbf{68.52$\pm$0.79} &
  \textbf{36.46$\pm$0.22} &
  38.00$\pm$0.28 &
  \textbf{36.60$\pm$0.37} &
  24.81$\pm$0.25 &
  26.28$\pm$0.24 &
  25.20$\pm$0.21 &
  0.91 \\
 &
  L-M &
  \textbf{75.62$\pm$0.10} &
  \textbf{72.34$\pm$0.82} &
  65.55$\pm$3.14 &
  31.44$\pm$0.30 &
  31.82$\pm$0.31 &
  31.72$\pm$0.23 &
  \textbf{30.69$\pm$0.26} &
  \textbf{31.01$\pm$00.29.} &
  \textbf{29.17$\pm$0.32} &
  2.89 \\
                        & L-L & 69.54$\pm$0.08 & 66.61$\pm$0.69 & 64.66$\pm$0.74 & 25.17$\pm$0.22 & 26.35$\pm$0.15 & 24.94$\pm$0.14 & 29.77$\pm$0.25  & 28.60$\pm$0.24 & 27.73$\pm$0.31 & 11.21 \\ \bottomrule
\end{tabular}%
\label{accuracy}
\vspace{-0.1in}
\end{table*}

\begin{figure*}[h]

	\centering
	\subfigure[Large-Small Model with IID]{
		\centering
		\includegraphics[width=0.25\textwidth]{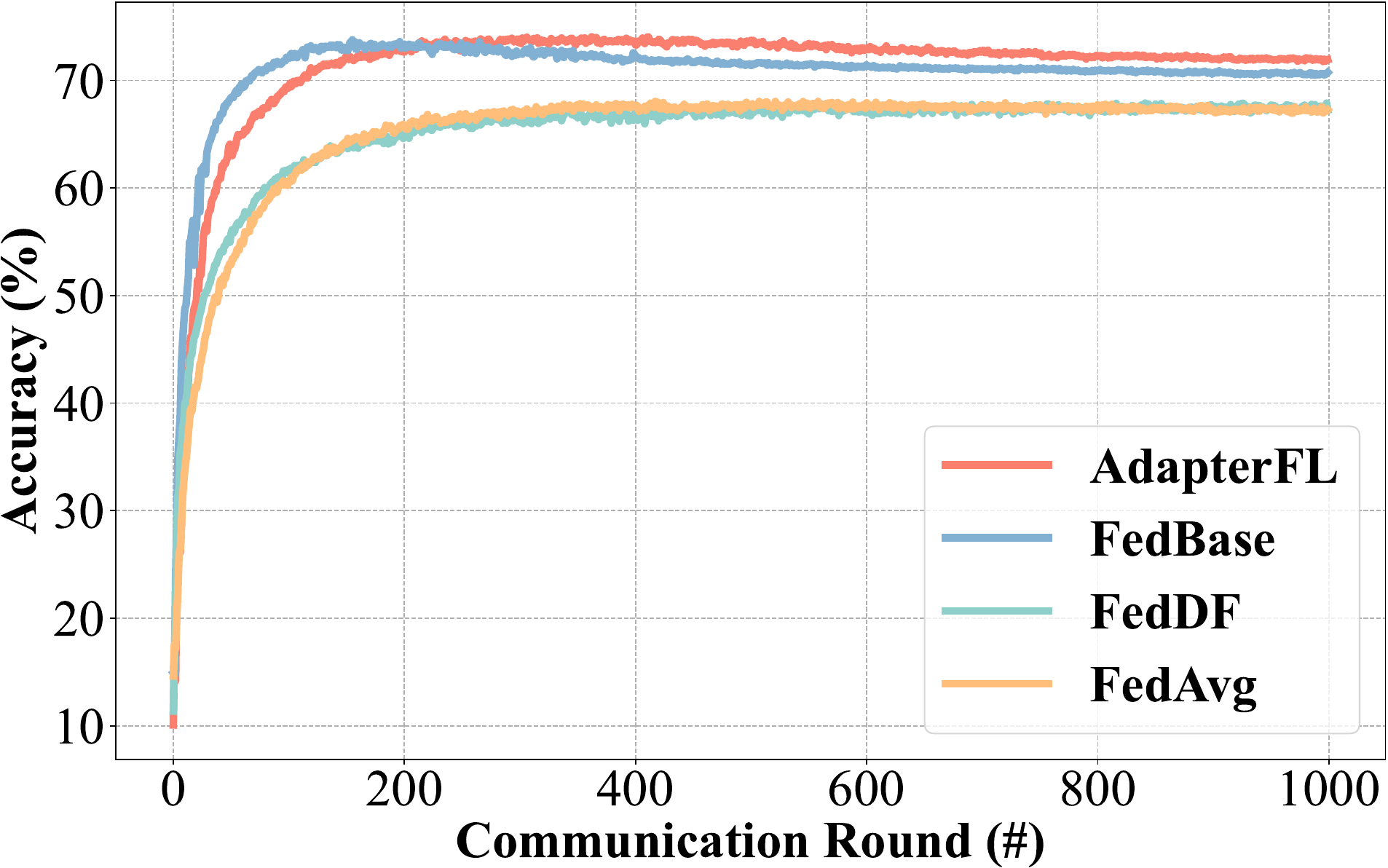}
		\label{fig:cifar-10-IID-ls}
	}
	\subfigure[Large-Medium Model with IID]{
		\centering
		\includegraphics[width=0.25\textwidth]{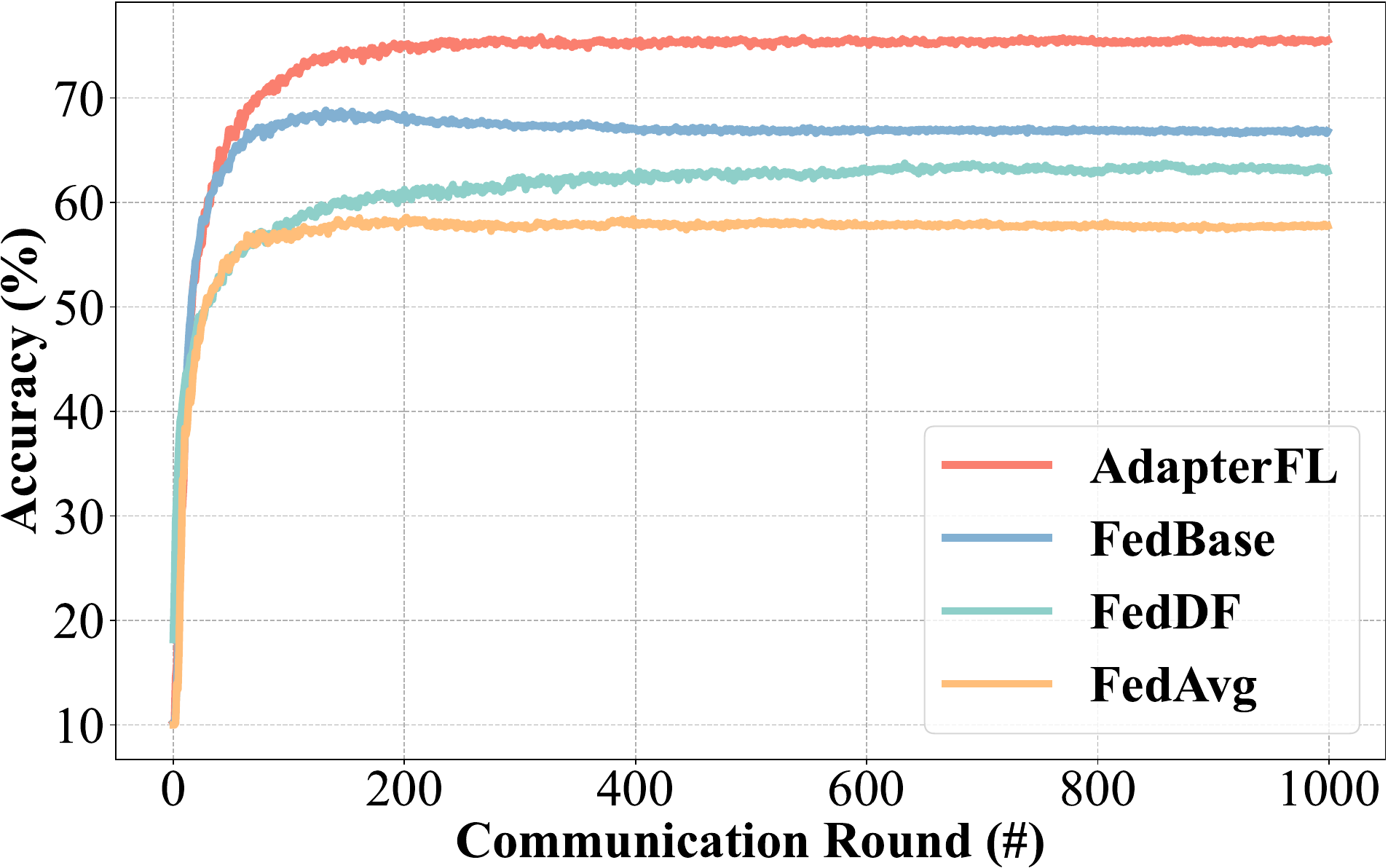}
		\label{fig:cifar-10-IID-lm}
	}
	\subfigure[Large-Large Model with IID]{
		\centering
		\includegraphics[width=0.25\textwidth]{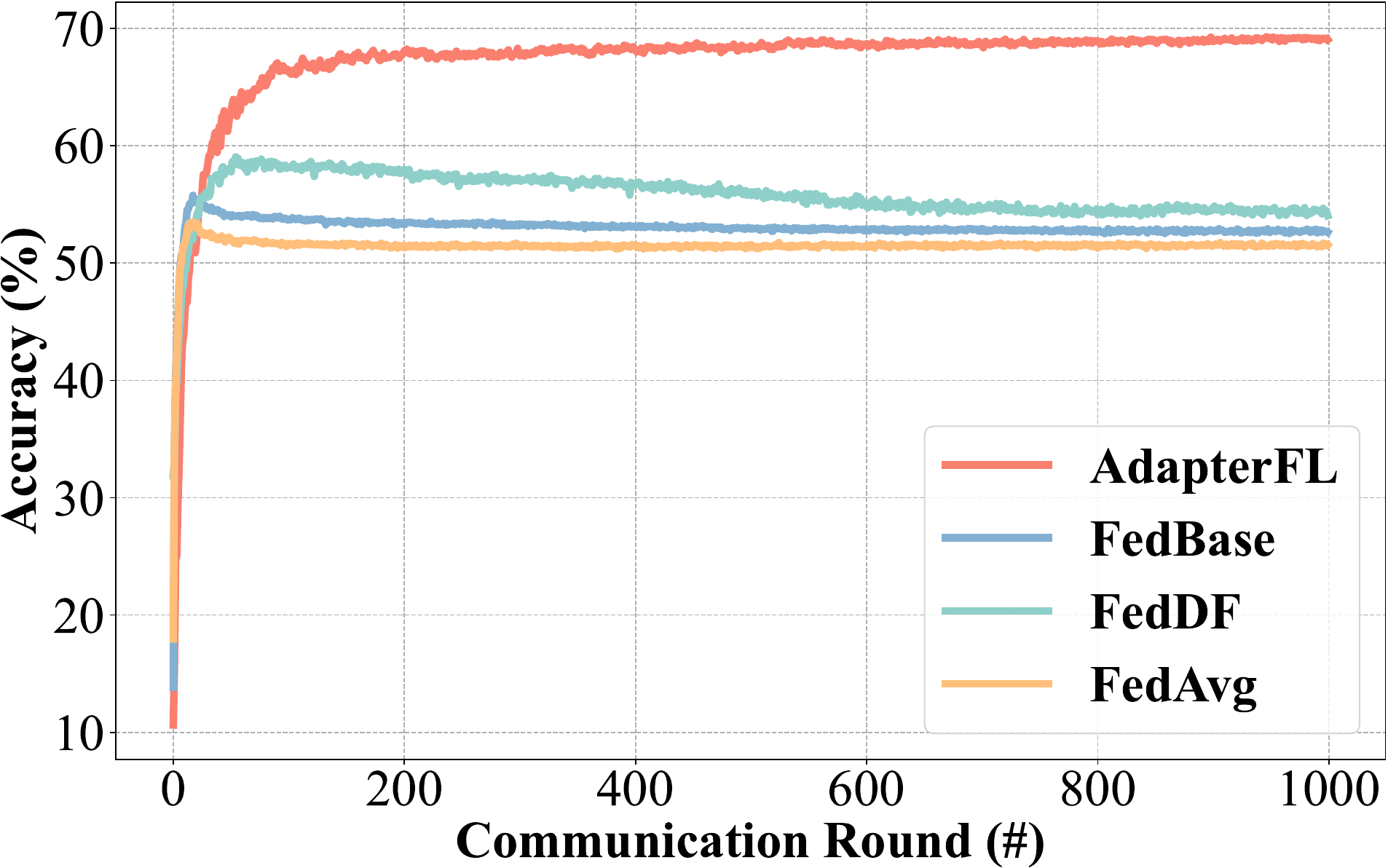}
		\label{fig:cifar-10-IID-ll}
	}
 	\centering
	\subfigure[Large-Small Model with $\beta = 0.6$ ]{
		\centering
		\includegraphics[width=0.25\textwidth]{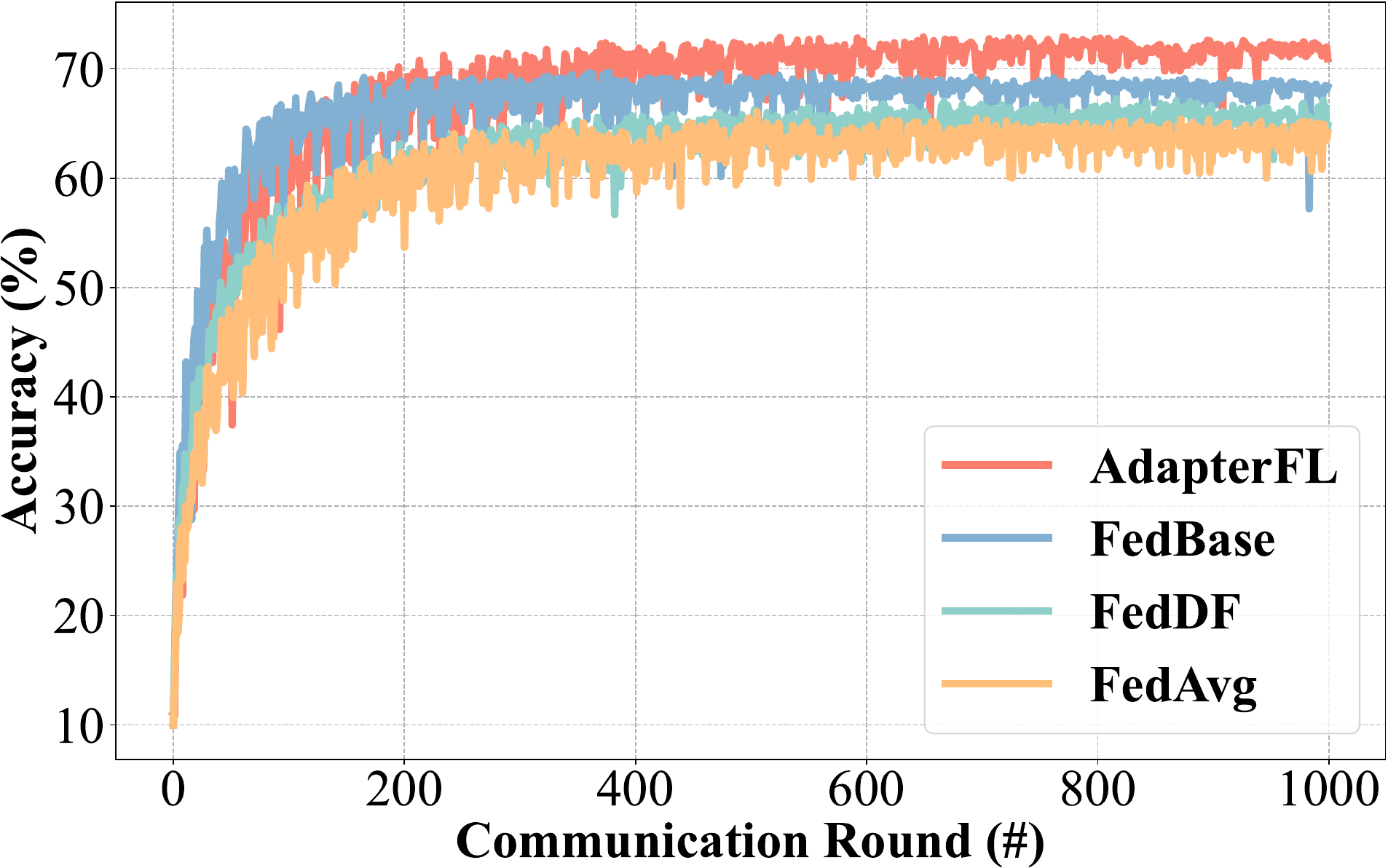}
		\label{fig:cifar-10-d0.6-ls}
	}
	\subfigure[Large-Medium Model with $\beta = 0.6$]{
		\centering
		\includegraphics[width=0.25\textwidth]{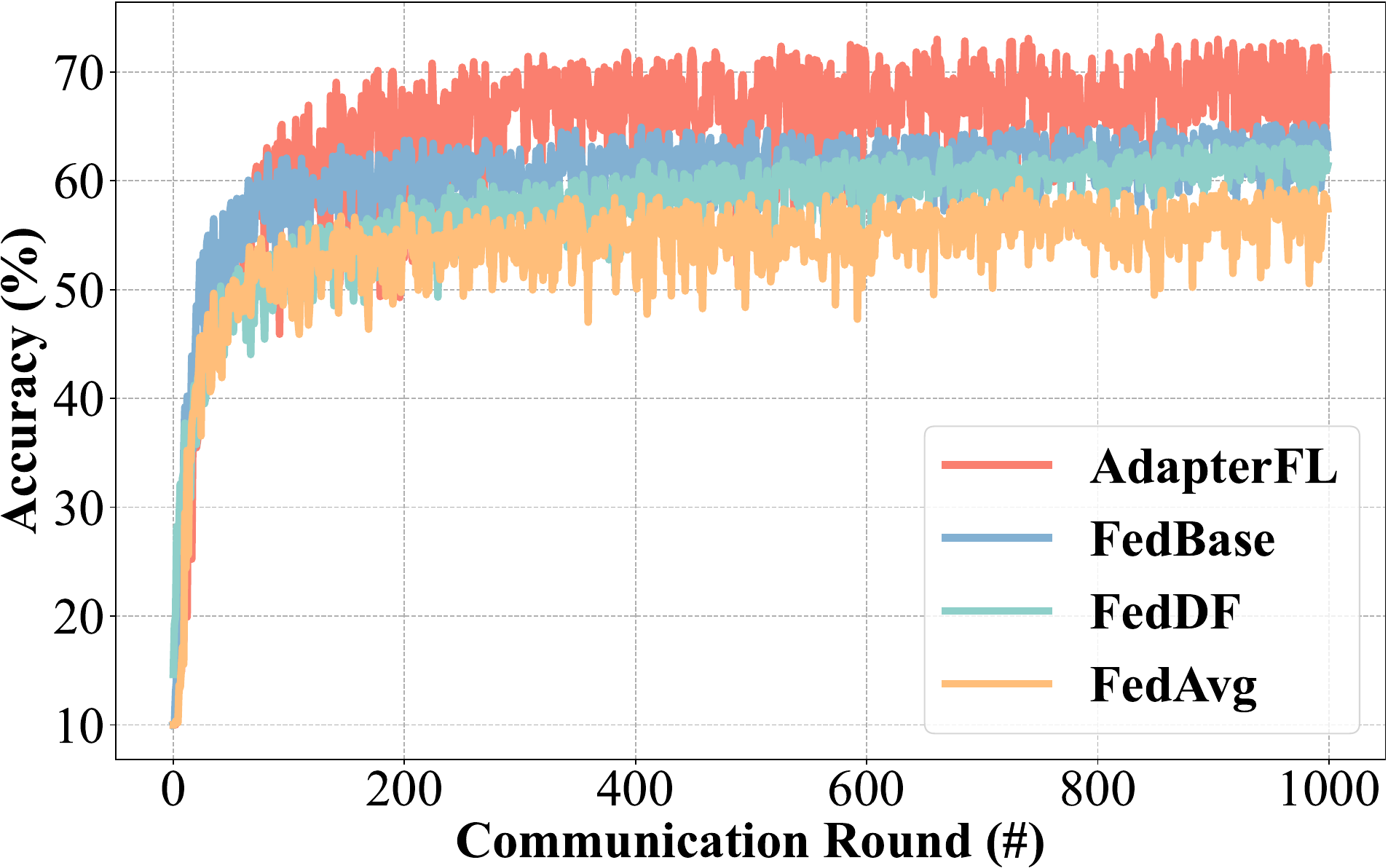}
		\label{fig:cifar-10-d0.6-lm}
	}\hspace{0.05in}
	\subfigure[Large-Large Model with $\beta = 0.6$]{
		\centering
		\includegraphics[width=0.25\textwidth]{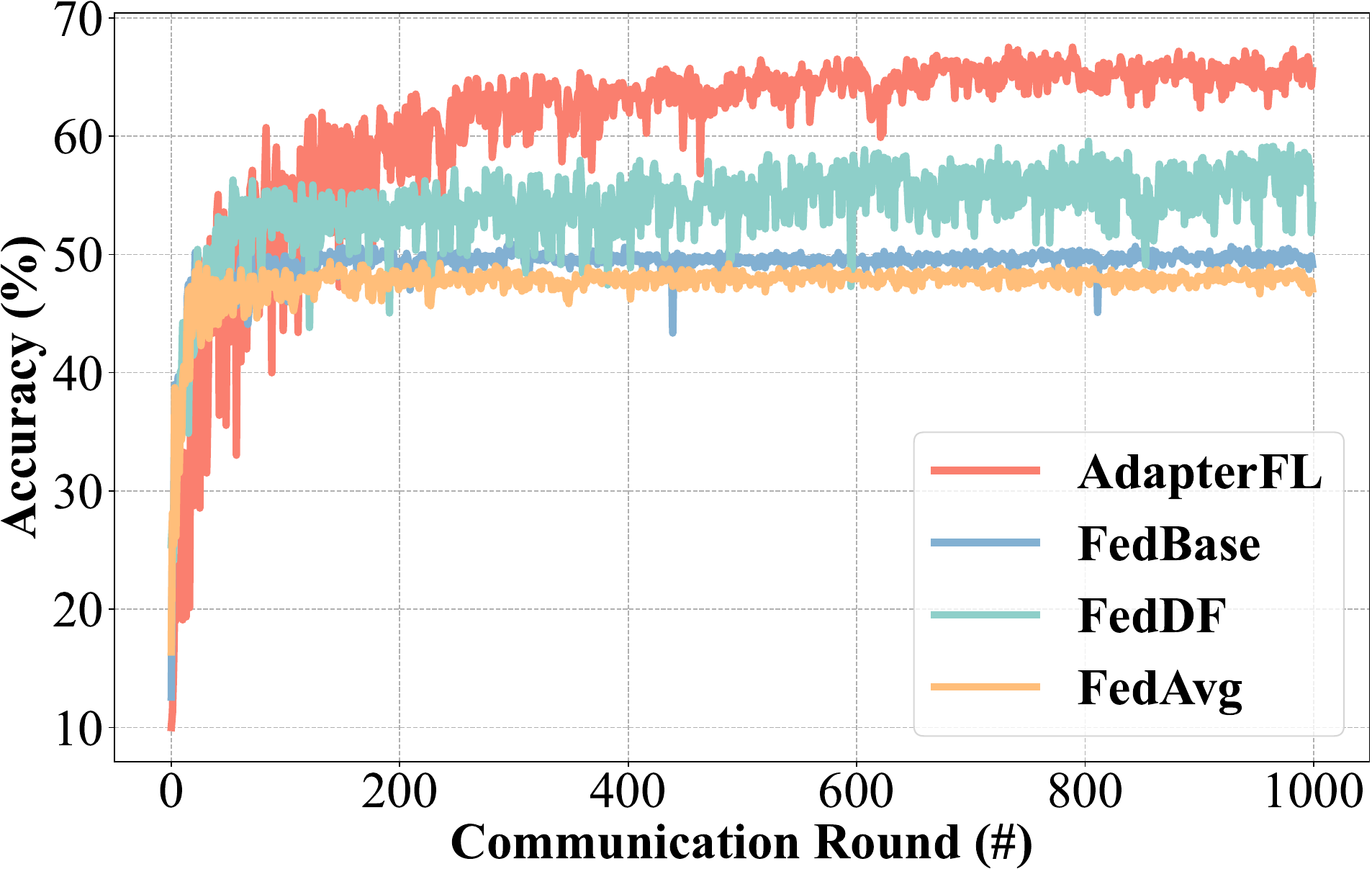}
		\label{fig:cifar-10-d0.6-ll}
	} 	
        \centering
	\subfigure[Large-Small Model with $\beta = 0.3$]{
		\centering
		\includegraphics[width=0.25\textwidth]{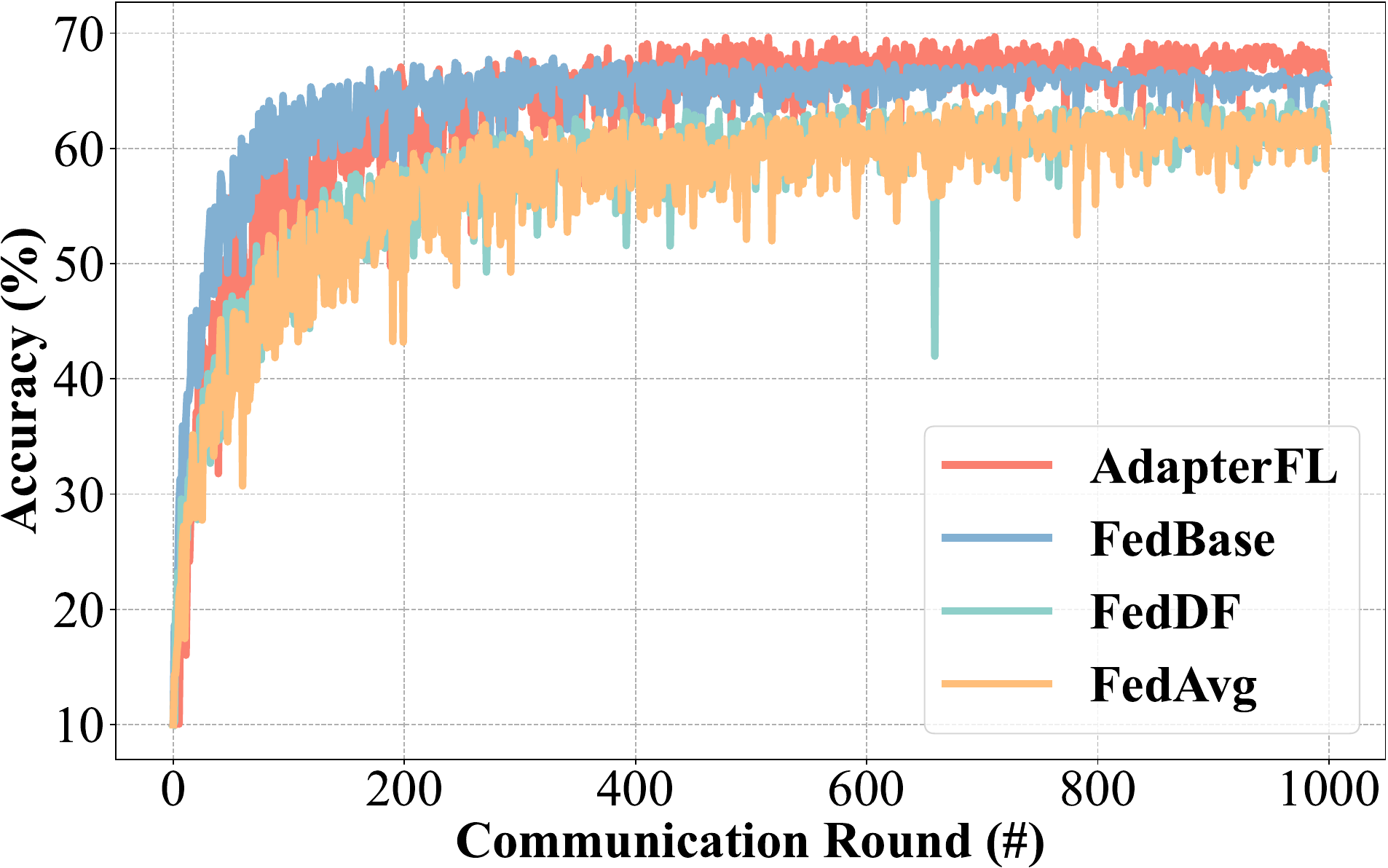}
		\label{fig:cifar-10-d0.3-ls}
	}
	\subfigure[Large-Medium Model with $\beta = 0.3$]{
		\centering
		\includegraphics[width=0.25\textwidth]{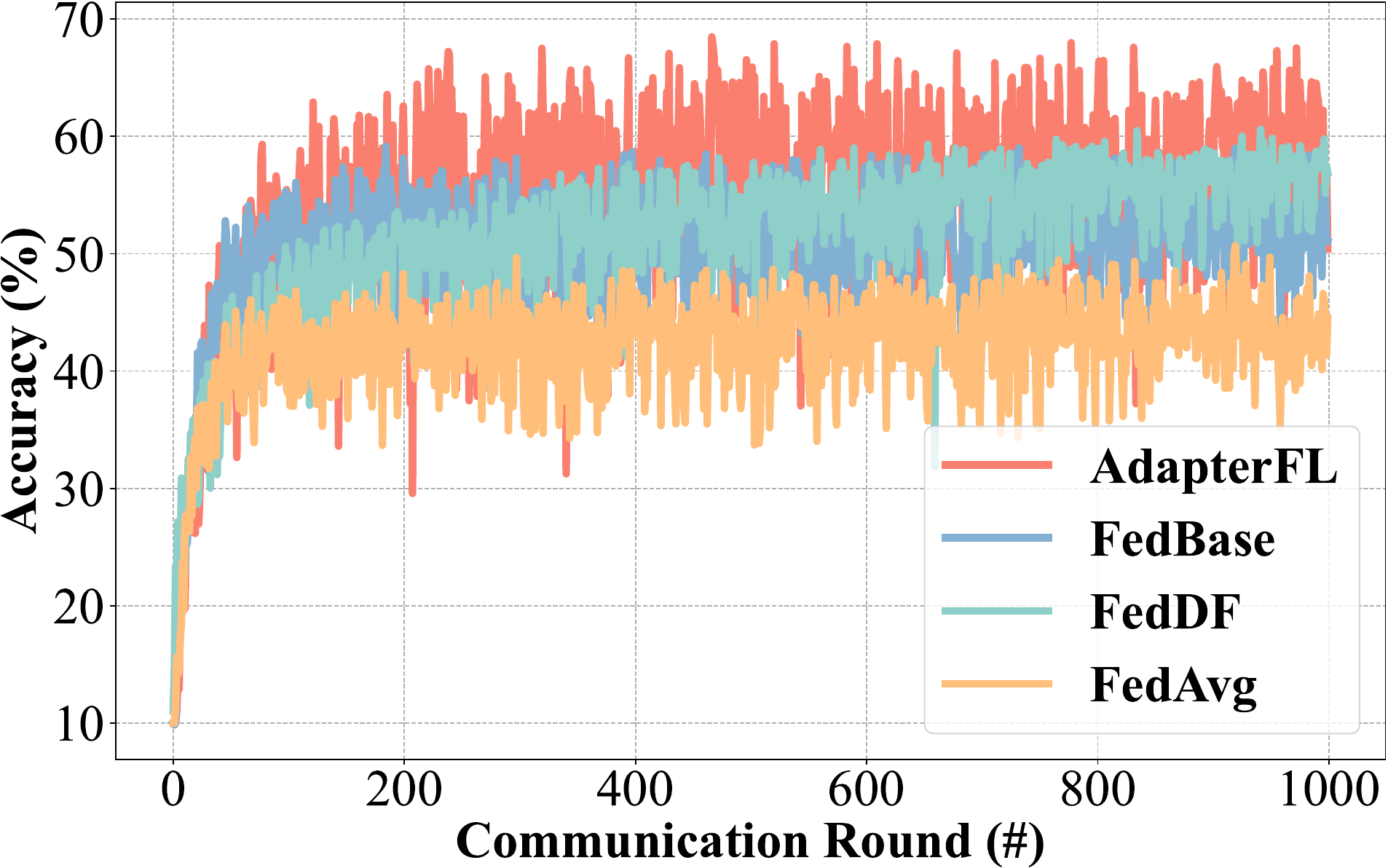}
		\label{fig:cifar-10-d0.3-lm}
	}
	\subfigure[Large-Large Model with $\beta = 0.3$]{
		\centering
		\includegraphics[width=0.25\textwidth]{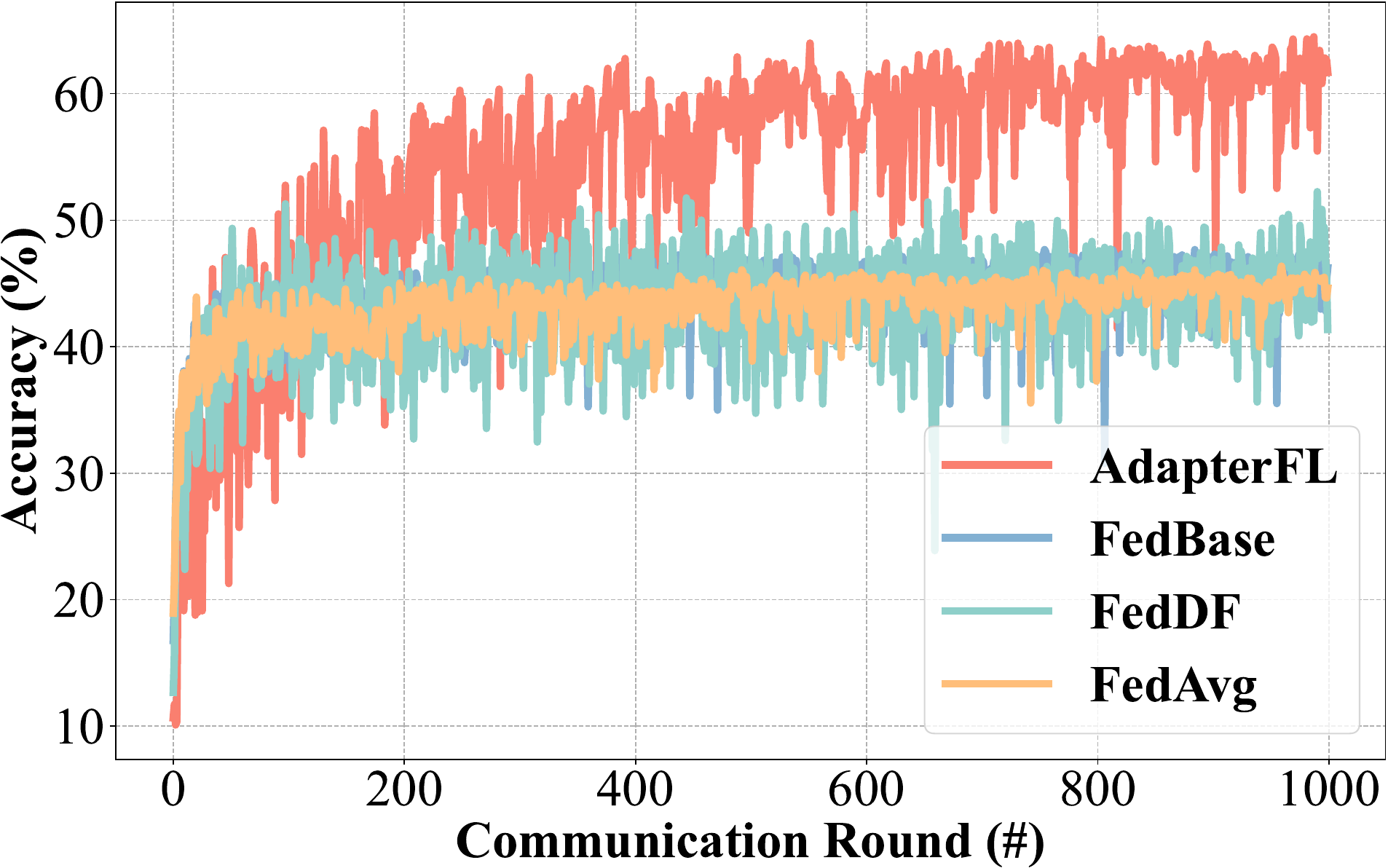}
		\label{fig:cifar-10-d0.3-ll}
	}
	\vspace{-0.05in}
	\caption{Learning curves of different FL methods of L-group on CIFAR-10.}
	\label{fig:learning_curve}
	\vspace{-0.05in}
\end{figure*}

\subsection{Performance Comparison (RQ1)}
Table~\ref{accuarcy_reassembly_models} compares the performance between AdapterFL and three existing methods using heterogeneous models on the CIFAR-10 dataset with three data distributions (IID, $\beta = 0.6$, and $\beta = 0.3$). In FedAvg and FedDF, S, M, and L represent three prototype models, e.g., L indicates the prototype model of ResNet18. In our method, we can obtain 3 groups of models, and each group contains 3 heterogeneous reassembled models composed of the same feature-extraction block. For example, the L-group includes 3 models: L-S, L-M, and L-L, and model L-M represents the reassembled model by combining the feature-extraction block of prototype model L and the device-adaptation block of prototype model M. We can observe that AdapterFL achieves the highest accuracy among all the mentioned methods in all levels of models by using the L-group. By the block-based aggregation, AdapterFL can outperform FedBase by $0.44\%, 7.23\%, 14.28\%  $, FedDF by $6.24\%, 12.06\%, 10.94\%$ and FedAvg by $5.98\%, 17.34\%, 16.33\%$ with three parameter levels of models of L-group on the IID data distribution. Compared to FedDF, which only improves the accuracy of the L and M model using additional global data by distilling, AdapterFL not only enhances the accuracy of model L-M and L-L, but it also works on model L-S without using the proxy dataset. Even in some cases, models with fewer parameters have higher accuracy. For different data distributions, AdapterFL can still perform better than other baselines. 

Table~\ref{accuarcy_reassembly_models} also shows the accuracy of different groups of models in our method.
By comparing the results of different groups, we can find performance differences between different groups of reassembled models. This is mainly because different feature-extraction blocks have different capabilities in extracting features, which is also the reason why the performance of the S-group is poor. We can observe that reassembled models composed of the feature-extraction block from the large prototype model outperform the model composed of the feature-extraction block from the small prototype model at the same parameter level, e.g., the model L-S outperforms model S-S by $7.61\%$ in FedBase and $6.24\%$ in AdapterFL. This indicates that the feature-extraction block of the large prototype model has more powerful feature extraction and knowledge transfer capabilities than the feature-extraction block of the small prototype model. 
The last column in Table~\ref{accuarcy_reassembly_models} lists the number of parameters of each model. We can calculate that the parameters are mainly concentrated on the device-adaptation block. Since the variation of the number of parameters in feature-extraction blocks and the addition of adapters, the number of parameters of reassembled models in AdapterFL is slightly
higher than that of corresponding prototype models. 
However, through the collaborative aggregation of feature-extraction blocks between heterogeneous models, AdapterFL can improve the overall performance of models by general feature extraction and knowledge transfer.

\begin{table}[]
\caption{Comparison of reassembly models on CIFAR-10 (S: MobileNetV2, M: ResNet18, L: Vgg16)}
\vspace{-0.1in}
\centering
\scriptsize
\begin{tabular}{@{}cc|ccc|c@{}}
\toprule
\multirow{2}{*}{Method}    & \multirow{2}{*}{Model} & \multicolumn{3}{c|}{Accuracy (\%)}                                 & \multirow{2}{*}{Params (M)} \\ \cmidrule(lr){3-5}
                         &     & IID         & $\beta=0.6$      & $\beta=0.3$      &       \\ \midrule
\multirow{3}{*}{FedAvg}    & S   & 64.61$\pm$0.10           & 63.01$\pm$1.10            & 53.98$\pm$1.67            & 2.25  \\
                           & M   & 64.91$\pm$0.18           & 66.19$\pm$0.14            & 58.91$\pm$0.22            & 11.17 \\
                           & L   & 66.40$\pm$0.08           & 63.56$\pm$0.17            & 59.71$\pm$0.16            & 33.65 \\ \midrule
\multirow{3}{*}{FedDF}     & S   & 71.60$\pm$0.10           & 69.21$\pm$0.48            & 66.49$\pm$0.56            & 2.25  \\
                           & M   & 69.64$\pm$0.09           & 67.72$\pm$0.29            & 65.19$\pm$0.35            & 11.17 \\
                           & L   & 69.82$\pm$0.13           & 68.29$\pm$0.33            & 62.46$\pm$0.30            & 33.65 \\ \midrule
\multirow{9}{*}{FedBase}   & S-S & 66.09$\pm$0.22           & 62.39$\pm$0.58            & 54.33$\pm$1.44            & 2.26  \\
                           & S-M & 61.54$\pm$0.18           & 60.18$\pm$0.53            & 52.40$\pm$1.51            & 10.58 \\
                           & S-L & 56.67$\pm$0.17           & 53.05$\pm$0.27            & 48.22$\pm$0.30            & 33.46 \\
                           & M-S & 72.31$\pm$0.13           & 69.12$\pm$0.63            & 62.31$\pm$1.55            & 2.89  \\
                           & M-M & 65.47$\pm$0.20           & 67.91$\pm$0.11            & 59.70$\pm$0.35            & 11.21 \\
                           & M-L & 66.86$\pm$0.12           & 62.41$\pm$0.17            & 57.18$\pm$0.25            & 34.10  \\
                           & L-S & 71.86$\pm$0.06           & 69.69$\pm$0.56            & 60.65$\pm$1.33            & 2.48  \\
                           & L-M & 71.10$\pm$0.10           & 72.95$\pm$0.12            & 69.32$\pm$0.18            & 10.79 \\
                           & L-L & 68.26$\pm$0.13           & 65.17$\pm$0.29            & 60.45$\pm$0.22            & 33.68 \\ \midrule
\multirow{9}{*}{AdapterFL} & S-S & 67.14$\pm$0.17           & 64.24$\pm$1.95            & 56.48$\pm$2.33            & 2.26  \\
                           & S-M & 66.50$\pm$0.09           & 66.02$\pm$0.34            & 60.47$\pm$0.86            & 10.58 \\
                           & S-L & 63.70$\pm$0.07           & 59.09$\pm$0.44            & 53.68$\pm$0.47            & 33.46 \\
                           & M-S & 75.63$\pm$0.14           & 73.14$\pm$1.28            & 65.04$\pm$2.14            & 2.89  \\
                           & M-M & 69.85$\pm$0.13           & 71.60$\pm$0.36            & 65.00$\pm$0.38            & 11.21 \\
                           & M-L & 71.20$\pm$0.08           & 67.86$\pm$0.55            & 63.17$\pm$0.72            & 34.10  \\
                           & L-S & 74.67$\pm$0.05           & 72.90$\pm$0.69            & 63.12$\pm$3.61            & 2.48  \\
                           & L-M & \textbf{75.30$\pm$0.08}  & \textbf{75.01$\pm$0.17}   & \textbf{70.75$\pm$0.28}   & 10.79 \\
                           & L-L & 73.07$\pm$0.06           & 71.19$\pm$0.40            & 64.54$\pm$0.36            & 33.68 \\ \bottomrule
\end{tabular}%
\label{accuary_different_models}
\vspace{-0.1in}
\end{table}

\subsection{Adaptivity Analysis (RQ2)}
Since Table~\ref{accuarcy_reassembly_models} shows that reassembled models of L-group can achieve the highest accuracy, we select the L-group as representatives of AdapterFL in the following experiments to prove the adaptivity. Experimental results of other groups under different data sets and data distributions will be shown in Appendix~\ref{additional}.

\subsubsection{Impacts of Datasets and client data distribution:} 
To prove the adaptivity of AdapterFL, we conducted experiments on 3 datasets (CIFAR-10, CIFAR-100, and TinyImageNet) with both IID and non-IID scenarios. For the non-IID scenario of each dataset, we set up two data heterogeneity scenarios (with $\beta = 0.6$ and $\beta = 0.3$, respectively). Table~\ref{accuracy} shows that our method can significantly improve the accuracy compared to the baselines on different datasets and data distributions. AdapterFL performs best even on the 200-class TinyImageNet data set and extreme non-IID data distribution.
The learning curves of different FL methods of L-group for the CIFAR-10 dataset can be seen in Figure~\ref{fig:learning_curve}.

\subsubsection{Impacts of Models:}

Comparing the number of parameters of models in Table~\ref{parameter} and Table~\ref{accuarcy_reassembly_models}, we can observe that the parameters of models in AdapterFL will increase slightly due to the addition of adapters. 
Therefore, to eliminate this impact and prove that the model does not limit our method, we selected three models with more parameters (MobileNetV2, ResNet18, and Vgg16~\cite{Vgg}) as small, medium, and large prototype models, respectively, and conducted experiments on the CIFAR-10 dataset in the IID scenarios. Table~\ref{accuary_different_models} shows that our method shows the best performance compared to other existing methods and the L-group can achieve the best results, which also illustrates that feature-extraction blocks from larger prototype models have more robust feature extraction and generalization capabilities. 

\begin{table}[h]
\caption{Accuracy comparison under different client resource ratios with the L-group on CIFAR-10}
\vspace{-0.1in}

\setlength\tabcolsep{3pt}
\centering
\scriptsize
\begin{tabular}{@{}cc|clccc@{}}
\toprule
\multirow{2}{*}{Method}    & \multirow{2}{*}{Model} & \multicolumn{5}{c}{Ratio of clients(Small, Mid, Large)}                          \\ \cmidrule(l){3-7} 
                         &     & 0.8/0.1/0.1 & \multicolumn{1}{c}{0.4/0.4/0.2} & 0.3/0.3/0.4         & 0.2/0.4/0.4 & 0.1/0.1/0.8         \\ \midrule
\multirow{3}{*}{FedBase} & L-S & 73.17$\pm$0.21  & 68.97$\pm$0.40                      & 73.34$\pm$0.21          & 73.11$\pm$0.18  & 73.02$\pm$0.10          \\
                         & L-M & 56.96$\pm$0.18  & 64.77$\pm$0.45                      & 69.61$\pm$0.17          & 70..49$\pm$0.11 & 72.03$\pm$0.16          \\
                         & L-L & 48.26$\pm$0.20  & 50.57$\pm$0.36                      & 61.76$\pm$0.16          & 62.80$\pm$0.18  & 69.69$\pm$0.17          \\ \midrule
\multirow{3}{*}{AdapterFL} & L-S                    & \textbf{73.24$\pm$0.19} & 72.32$\pm$0.45 & 74.33$\pm$0.11 & \textbf{75.77$\pm$0.18} & 76.83$\pm$0.06 \\
                         & L-M & 72.83$\pm$0.20  & \textbf{72.34$\pm$0.82}             & \textbf{77.06$\pm$0.12} & 75.74$\pm$0.13  & \textbf{79.32$\pm$0.09} \\
                         & L-L & 68.46$\pm$0.14  & 66.61$\pm$0.69                      & 72.35$\pm$0.05          & 70.72$\pm$0.08  & 72.16$\pm$0.07          \\ \bottomrule
\end{tabular}%
\vspace{-0.1in}
\label{accuracy_different_ratio}
\end{table}

\subsection{Scalability Analysis (RQ3)}
\subsubsection{Imapcts of clients resource distribution ratios}
To evaluate the scalability of AdapterFL, we conducted experiments at different client resource ratios and compared AdapterFL with different baselines, as shown in Table~\ref{accuracy_different_ratio}. As we thought, the model's overall performance will improve as the proportion of resource-constrained clients decreases. Not only will the accuracy of large models be improved, but that of small models will also be significantly improved. This is because as the proportion of clients with sufficient resources gradually increases, large models can be better trained and their feature-extraction block can more fully utilize its capabilities to extract the common feature.



\begin{table}[h]
\caption{Accuracy comparison under different activated client ratios with the L-group on CIFAR-10}
\vspace{-0.1in}
\centering
\scriptsize
\begin{tabular}{@{}cc|cccc@{}}
\toprule
\multirow{2}{*}{Method}    & \multirow{2}{*}{Model} & \multicolumn{4}{c}{Ratios of activated clients}                                       \\ \cmidrule(l){3-6} 
 &     & 0.05       & 0.1        & 0.5        & 1          \\ \midrule
\multirow{3}{*}{FedBase}   & L-S                    & 73.49$\pm$0.19          & 73.38$\pm$0.29          & 72.55$\pm$0.07          & 72.17$\pm$0.07          \\
 & L-M & 69.47$\pm$0.14 & 68.39$\pm$0.21 & 67.45$\pm$0.10 & 67.23$\pm$0.15 \\
 & L-L & 55.84$\pm$0.20 & 55.28$\pm$0.22 & 53.85$\pm$0.23 & 54.76$\pm$0.20 \\ \midrule
\multirow{3}{*}{AdapterFL} & L-S                    & 75.11$\pm$0.15          & 73.82$\pm$0.11          & 73.62$\pm$0.08          & \textbf{73.14$\pm$0.09} \\
                           & L-M                    & \textbf{76.46$\pm$0.12} & \textbf{75.62$\pm$0.10} & \textbf{74.09$\pm$0.16} & 73.09$\pm$0.13          \\
 & L-L & 73.01$\pm$0.06 & 69.54$\pm$0.08 & 66.42$\pm$0.06 & 65.35$\pm$0.02 \\ \bottomrule
\end{tabular}%
\vspace{-0.1in}
\label{accuracy_different_active}
\end{table}

\begin{table}[h]
\caption{Accuracy comparison under different total numbers of clients with the L-group on CIFAR-10}
\vspace{-0.1in}
\centering
\scriptsize
\begin{tabular}{@{}cc|cccc@{}}
\toprule
\multirow{2}{*}{Method}    & \multirow{2}{*}{Model} & \multicolumn{4}{c}{Number of clients}                                                 \\ \cmidrule(l){3-6} 
                           &                        & 50                  & 100                 & 200                 & 500                 \\ \midrule
\multirow{3}{*}{FedBase}   & L-S                    & 76.29$\pm$0.15          & 73.38$\pm$0.29          & 70.58$\pm$0.19          & 64.98$\pm$0.20          \\
                           & L-M                    & 70.98$\pm$0.19          & 68.39$\pm$0.21          & 64.50$\pm$0.23          & 58.24$\pm$0.13          \\
                           & L-L                    & 57.07$\pm$0.09          & 55.28$\pm$0.22          & 51. 86$\pm$0.10         & 49.04$\pm$0.10          \\ \midrule
\multirow{3}{*}{AdapterFL} & L-S                    & 77.72$\pm$0.10          & 73.82$\pm$0.11          & 70.79$\pm$0.08          & \textbf{66.73$\pm$0.12} \\
                           & L-M                    & \textbf{79.09$\pm$0.10} & \textbf{75.62$\pm$0.10} & \textbf{71.86$\pm$0.18} & 65.04$\pm$0.13          \\
                           & L-L                    & 73.97$\pm$0.09          & 69.54$\pm$0.08          & 64.26$\pm$0.07          & 58.89$\pm$0..10         \\ \bottomrule
\end{tabular}%
\vspace{-0.1in}
\label{accuracy_different_clients}
\end{table}

\subsubsection{Impact of number of clients} 
We conducted experiments with different ratios of activated clients and total numbers of clients on the CIFAR-10 dataset in the IID scenario. Table~\ref{accuracy_different_clients} shows the experimental results of our method with the total number of clients $N = 50, 100, 200, 500$, and Table~\ref{accuracy_different_active} shows the results with the total number of clients $\alpha = 0.05, 0.1, 0.5, 1.0$.
As the ratios of activated clients and the overall number of clients increase, the overall performance of models will decrease. However, AdapterFL can still perform better than the baseline under various client settings.


\section{Conclusion}
\label{Conclusion}
In this paper, we first introduced the resource-constrained issue in actual scenarios and then discussed the challenge of FL: model heterogeneity. Then, We proposed AdapterFL, a heterogeneous FL framework based on model partition and reassembly for model heterogeneity. Our method does not need additional datasets and can still work on models with different architectures. Compared with state-of-the-art FL frameworks, AdpaterFL improves model accuracy significantly in the resource-limited scenario. We also proved the adaptivity and scalability of our method through experiments.  


\bibliographystyle{IEEEtran}
\bibliography{IEEEexample}


\clearpage
\appendix
\section{APPENDIX}
\subsection{Additional Experiments}
\label{additional}
Table~\ref{cifar100_cnn} and Table~\ref{TinyImageNet} shows the performance comparison between AdapterFL and three existing methods on CIFAR-10 and TinyImageNet respectively with three data distributions (IID, $\beta$ = 0.6, $\beta$ = 0.3). Table~\ref{cifar100_vgg} shows performance variations using MobileNetV2, ResNet18, and Vgg16 as prototype models on the CIFAR-100 dataset. Under these different settings, we can see that our method still works well, and the L-group model can achieve the best accuracy.
\begin{table}[h]
\caption{Comparison of reassembled models on CIFAR-100 }
\vspace{-0.1in}
\centering
\scriptsize
\begin{tabular}{@{}cc|ccc|c@{}}
\toprule
\multirow{2}{*}{Method}    & \multirow{2}{*}{Model} & \multicolumn{3}{c|}{Accuracy (\%)}                                 & \multirow{2}{*}{Params (M)} \\ \cmidrule(lr){3-5}
                         &     & IID         & $\beta=0.6$       & $\beta=0.3$      &       \\ \midrule
\multirow{3}{*}{FedAvg}    & S   & 23.41$\pm$0.09           & 25.20$\pm$0.20            & 25.15$\pm$0.12            & 0.21  \\
                           & M   & 21.81$\pm$0.21           & 24.68$\pm$0.14            & 23.36$\pm$0.20            & 2.25  \\
                           & L   & 20.90$\pm$0.20           & 19.59$\pm$0.43            & 18.26$\pm$0.22            & 11.17 \\ \midrule
\multirow{3}{*}{FedDF}     & S   & 26.94$\pm$0.13           & 29.56$\pm$0.17            & 28.84$\pm$0.28            & 0.21  \\
                           & M   & 25.44$\pm$0.16           & 25.90$\pm$0.24            & 25.57$\pm$0.11            & 2.25  \\
                           & L   & 22.44$\pm$0.23           & 21.92$\pm$0.26            & 21.08$\pm$0.53            & 11.17 \\ \midrule
\multirow{9}{*}{FedBase}   & S-S & 19.22$\pm$0.26           & 22.60$\pm$0.21            & 22.85$\pm$0.30            & 0.22  \\
                           & S-M & 21.39$\pm$0.22           & 21.77$\pm$0.19            & 20.58$\pm$0.22            & 2.20  \\
                           & S-L & 22.25$\pm$0.18           & 20.71$\pm$0.08            & 19.01$\pm$0.13            & 10.52 \\
                           & M-S & 28.01$\pm$0.14           & 28.97$\pm$0.21            & 28.57$\pm$0.25            & 0.28  \\
                           & M-M & 22.41$\pm$0.18           & 23.69$\pm$0.24            & 23.76$\pm$0..07           & 2.26  \\
                           & M-L & 18.80$\pm$0.07           & 20.52$\pm$0.07            & 18.38$\pm$0.05            & 10.58 \\
                           & L-S & 35.80$\pm$0.22           & \textbf{38.84$\pm$0.24}   & 34.91$\pm$0.34            & 0.91  \\
                           & L-M & 24.18$\pm$0.17           & 24.80$\pm$0.28            & 23.73$\pm$0.24            & 2.89  \\
                           & L-L & 20.44$\pm$0.17           & 19.55$\pm$0.33            & 18.45$\pm$0.14            & 11.21 \\ \midrule
\multirow{9}{*}{AdapterFL} & S-S & 22.36$\pm$0.31           & 24.10$\pm$0.22            & 24.23$\pm$0.26            & 0.22  \\
                           & S-M & 21.73$\pm$0.21           & 22.04$\pm$0.18            & 21.51$\pm$0.16            & 2.20  \\
                           & S-L & 22.73$\pm$0.12           & 22.54$\pm$0.05            & 20.82$\pm$0.16            & 10.52 \\
                           & M-S & 29.51$\pm$0.17           & 32.61$\pm$0.29            & 30.98$\pm$0..21           & 0.28  \\
                           & M-M & 26.16$\pm$0.23           & 29.41$\pm$0.16            & 27.67$\pm$0.14            & 2.26  \\
                           & M-L & 26.71$\pm$0.07           & 28.31$\pm$0.10            & 25.86$\pm$0.15            & 10.58 \\
                           & L-S & \textbf{36.46$\pm$0.22}  & 38.00$\pm$0.28            & \textbf{36.60$\pm$0.37}   & 0.91  \\
                           & L-M & 31.44$\pm$0.30           & 31.82$\pm$0.31            & 31.72$\pm$0.23            & 2.89  \\
                           & L-L & 25.17$\pm$0.22           & 26.35$\pm$0.15            & 24.94$\pm$0.14            & 11.21 \\ \bottomrule
\end{tabular}%
\label{cifar100_cnn}
\end{table}

\begin{table}[b]
\caption{Comparison of reassembled models on TinyImageNet }
\centering
\vspace{-0.1in}
\scriptsize
\begin{tabular}{@{}cc|ccc|c@{}}
\toprule
\multirow{2}{*}{Method}    & \multirow{2}{*}{Model} & \multicolumn{3}{c|}{Accuracy (\%)}                                 & \multirow{2}{*}{Params (M)} \\ \cmidrule(lr){3-5}
                         &     & IID         & $\beta=0.6$       & $\beta=0.3$       &       \\ \midrule
\multirow{3}{*}{FedAvg}    & S   & 11.87$\pm$0.10           & 13.95$\pm$0.20            & 15.38$\pm$0.14            & 0.21  \\
                           & M   & 23.23$\pm$0.27           & 23.67.0.22            & 22.72$\pm$0.25            & 2.25  \\
                           & L   & 27.57$\pm$0.23           & 26.25$\pm$0.43            & 24.69$\pm$0.44            & 11.17 \\ \midrule
\multirow{3}{*}{FedDF}     & S   & 19.69$\pm$0.22           & 20.73$\pm$0.27            & 21.47$\pm$0.24            & 0.21  \\
                           & M   & 30.51$\pm$0.18           & 29.99$\pm$0.17            & 30.16$\pm$0.24            & 2.25  \\
                           & L   & 26.24$\pm$00.30          & 24.18$\pm$0.25            & 22.29$\pm$0.40            & 11.17 \\ \midrule
\multirow{9}{*}{FedBase}   & S-S & 8.13$\pm$0.10            & 10.41$\pm$0.28            & 12.59$\pm$0.19            & 0.22  \\
                           & S-M & 23.28$\pm$0.27           & 23.37$\pm$0.28            & 22.18$\pm$0.18            & 2.20   \\
                           & S-L & 26.95$\pm$0.11           & 25.68$\pm$0.39            & 24.90$\pm$0.29                     & 10.52 \\
                           & M-S & 18.56$\pm$0.22           & 21.02$\pm$0.17            & 22.75$\pm$0.13            & 0.28  \\
                           & M-M & 22.86$\pm$0.17           & 24.37$\pm$0.37            & 24.38$\pm$0.38            & 2.26  \\
                           & M-L & 18.55$\pm$0.16           & 18.17$\pm$0.28            & 17.71$\pm$0.15            & 10.58 \\
                           & L-S & 22.67$\pm$0.22           & 24$\pm$20$\pm$0.18            & 23.96$\pm$0.23            & 0.91  \\
                           & L-M & 19.70$\pm$0.23           & 21.81$\pm$0.28            & 20.77$\pm$0.22            & 2.89  \\
                           & L-L & 29.24$\pm$0.13           & 27.60$\pm$0.32            & 26.43$\pm$0.48            & 11.21 \\ \midrule
\multirow{9}{*}{AdapterFL} & S-S & 12.71$\pm$0.14           & 13.27$\pm$0.22            & 14.52$\pm$0.26            & 0.22  \\
                           & S-M & 25.31$\pm$0.21           & 25.57$\pm$0.37            & 24.03$\pm$0.35            & 2.20   \\
                           & S-L & 25.35$\pm$0.36           & 23.77$\pm$0.16            & 22.91$\pm$0.14            & 10.52 \\
                           & M-S & 24.48$\pm$0.13           & 23.47$\pm$0.28            & 23.67$\pm$0.29            & 0.28  \\
                           & M-M & 26.75$\pm$0.13           & 25.63$\pm$0.33            & 24.99$\pm$0.27            & 2.26  \\
                           & M-L & 23.11$\pm$0.20           & 22.35$\pm$0.11            & 21.89$\pm$0.10            & 10.58 \\
                           & L-S & 24.81$\pm$0.25           & 26.28$\pm$0.24            & 25.20$\pm$0.21            & 0.91  \\
                        & L-M                    & \textbf{30.69$\pm$0.26} & \textbf{31.01$\pm$00.29.} & \textbf{29.17$\pm$0.32} & 2.89                        \\
                           & L-L & 29.77$\pm$0.25           & 28.60$\pm$0.24            & 27.73$\pm$0.31            & 11.21 \\ \bottomrule
\end{tabular}%
\vspace{-0.1in}
\label{TinyImageNet}
\end{table}

\begin{table}[b]
\caption{Comparison of reassembly models on CIFAR-100 (S: MobileNetV2, M: ResNet18, L: Vgg16)}
\centering
\vspace{-0.1in}
\scriptsize
\begin{tabular}{@{}cc|ccc|c@{}}
\toprule
\multirow{2}{*}{Method}    & \multirow{2}{*}{Model} & \multicolumn{3}{c|}{Accuracy (\%)}                                 & \multirow{2}{*}{Params (M)} \\ \cmidrule(lr){3-5}
                         &     & IID         & $\beta=0.6$       &$\beta=0.3$      &       \\ \midrule
\multirow{3}{*}{FedAvg}    & S   & 28.35$\pm$0.13           & 28.99$\pm$0.15            & 28.36$\pm$0.14            & 2.25  \\
                           & M   & 28.66$\pm$0.24           & 28.60$\pm$0.13            & 28.09$\pm$0.14            & 11.17 \\
                           & L   & 23.76$\pm$0.14           & 23.43$\pm$0.20            & 22.23$\pm$0.22            & 33.65 \\ \midrule
\multirow{3}{*}{FedDF}     & S   & 29.80$\pm$0.18           & 32.04$\pm$0.21            & 31.73$\pm$0.24            & 2.25  \\
                           & M   & 29.82$\pm$0.23           & 29.75$\pm$0.46            & 29.85$\pm$0.55            & 11.17 \\
                           & L   & 25.68$\pm$0.32           & 25.15$\pm$0.55            & 25.34$\pm$0.17            & 33.65 \\ \midrule
\multirow{9}{*}{FedBase}   & S-S & 27.68$\pm$0.16           & 29.35$\pm$0.24            & 27.77$\pm$0.25            & 2.26  \\
                           & S-M & 26.25$\pm$0.20           & 28.07$\pm$0.07            & 27.71$\pm$0.09            & 10.58 \\
                           & S-L & 17.41$\pm$0.19           & 16.80$\pm$0.10            & 17.16$\pm$0.12            & 33.46 \\
                           & M-S & 29.45$\pm$0.29           & 28.59$\pm$0.19            & 28.29$\pm$0.50            & 2.89  \\
                           & M-M & 29.59$\pm$0.22           & 30.03$\pm$0.12            & 28.47$\pm$0.14            & 11.21 \\
                           & M-L & 22.80$\pm$0.17           & 22.12$\pm$0.11            & 20.69$\pm$0.30            & 34.1  \\
                           & L-S & 30.65$\pm$0.15           & 31.27$\pm$0.12            & 30.76$\pm$0.08            & 2.48  \\
                           & L-M & 35.86$\pm$0.16           & 38.22$\pm$0.08            & 37.48$\pm$0.11            & 10.79 \\
                           & L-L & 24.36$\pm$0.22           & 22.60$\pm$0.13            & 21.87$\pm$0.10            & 33.68 \\ \midrule
\multirow{9}{*}{AdapterFL} & S-S & 28.05$\pm$0.19           & 30.05$\pm$0.15            & 29.21$\pm$0.01            & 2.26  \\
                           & S-M & 29.72$\pm$0.08           & 32.22$\pm$0.10            & 33.00$\pm$0.10            & 10.58 \\
                           & S-L & 20.08$\pm$0.06           & 19.25$\pm$0.11            & 19.75$\pm$0.07            & 33.46 \\
                           & M-S & 33.30$\pm$0.15           & 32.26$\pm$0.21            & 31.85$\pm$0.27            & 2.89  \\
                           & M-M & 32.37$\pm$0.24           & 33.90$\pm$0.08            & 32.55$\pm$0.15            & 11.21 \\
                           & M-L & 25.20$\pm$0.09           & 24.24$\pm$0.08            & 22.62$\pm$0.15            & 34.1  \\
                           & L-S & 32.90$\pm$0.24           & 35.73$\pm$0.16            & 34.84$\pm$0.13            & 2.48  \\
                           & L-M & \textbf{39.01$\pm$0.14}  & \textbf{41.90$\pm$0.17}   & \textbf{40.82$\pm$0.08}   & 10.79 \\
                           & L-L & 26.58$\pm$0.12           & 26.48$\pm$0.09            & 24.30$\pm$0.10            & 33.68 \\ \bottomrule
\end{tabular}%
\vspace{-0.1in}
\label{cifar100_vgg}
\end{table}

\end{document}